\documentclass[11pt]{article}
\usepackage[preprint]{acl}

\usepackage[utf8]{inputenc} 
\usepackage[T1]{fontenc}    
\usepackage{hyperref}       
\usepackage{url}            
\usepackage{booktabs}       
\usepackage{amsfonts}       
\usepackage{nicefrac}       
\usepackage{microtype}      
\usepackage{xcolor}         
\usepackage{newfloat}
\usepackage{listings}
\usepackage{times}
\usepackage{latexsym}

\usepackage[T1]{fontenc}

\usepackage[utf8]{inputenc}

\usepackage{microtype}
\usepackage{graphicx} 
\usepackage{booktabs}
\usepackage{multirow}
\usepackage{morefloats}

\usepackage{latexsym}

\usepackage{inconsolata}
\usepackage{amsfonts}
\usepackage{nameref}
\usepackage{amsmath}
\usepackage{cleveref}
\usepackage{caption}
\usepackage{subcaption}

\title{How Many Ratings per Item are Necessary\\for Reliable Significance Testing?}

\author{
 \textbf{Christopher M. Homan\textsuperscript{1}},
 \textbf{Flip Korn\textsuperscript{2}},
 \textbf{Deepak Pandita\textsuperscript{1}},
 \textbf{Chris Welty\textsuperscript{2}}
\\
\\
 \textsuperscript{1}Rochester Institute of Technology,
 \textsuperscript{2}Google Research
\\
 \small{
   \texttt{cmh@cs.rit.edu, flip@google.com, deepak@mail.rit.edu, cawelty@gmail.com} 
 }
}

\newcommand{\pv}{\textit{p}-value}
\newcommand{\pvs}{\textit{p}-values}

\newcommand{\mae}{$\Gamma_{\rm MAE}$}
\newcommand{\wins}{$\Gamma_{\rm Wins}$}

\newcommand{\hide}[1]{}

\newcommand{\mparams}{\mathbf{\theta}}
\newcommand{\itemdist}{P_{items}}
\newcommand{\respdist}{P_{responses}}
\newcommand{\normdist}{\mathcal{N}}

\newlength\myindent
\extrafloats{300}

\usepackage{bibentry}

\begin{document}

\date{}
\maketitle

\begin{abstract}
 A cornerstone of machine learning evaluation is the (often hidden) assumption that model and human responses are reliable enough to evaluate models against unitary, authoritative, ``gold standard'' data, via simple metrics such as accuracy, precision, and recall. The generative AI revolution would seem to explode this assumption, given the critical role stochastic inference plays. Yet, in spite of public demand for more transparency in AI---along with strong evidence that humans are unreliable judges---estimates of model reliability are conventionally based on, at most, a few output responses per input item. We adapt a method, previously used to evaluate the reliability of various metrics and estimators for machine learning evaluation, to determine whether an (existing or planned) dataset has enough responses per item to assure reliable null hypothesis statistical testing. We show that, for many common metrics, collecting even 5-10 responses per item (from each model and team of human evaluators) is not sufficient. We apply our methods to several of the very few extant gold standard test sets with multiple disaggregated responses per item and show that even these datasets lack enough responses per item. We show how our methods can help AI researchers make better decisions about how to collect data for AI evaluation.
 
\end{abstract}

\section{Introduction}

Arguably, the two central questions of experimental design are:
\emph{What degree of detection capability must the study possess to ensure that a genuine effect, if present, is measured?}
and
\emph{How reliably can we predict the same outcome in future trials, given the observed evidence?}
Here, \textbf{power analysis (PA)}~\cite{bausell2002power} helps to answer the first question
by controlling for false-negatives,
and \textbf{null hypothesis statistical tests} (NHSTs) -- or, in some cases, {confidence intervals} (CIs) -- address the second
by controlling for false-positives.

For AI evaluation, nearly all existing implementations of these fundamental tools for capturing experimental reproducibility measure only the variation of the inputs.
\emph{Yet they fail to capture the variance of the \textbf{output responses}---model or human---associated with each test input item.}

On the model side, response variance can come from stochastic inference, which is responsible for the creative power of foundation models, such as LLMs.  
It can also come from race conditions~\cite{DBLP:journals/corr/abs-2408-05148}, mixtures of experts~\cite{shazeer2017}, Monte Carlo dropout~\cite{gal16dropout}, and ensembling~\cite{balaji17ensembles}.

On the human side, annotation and feedback continue to play a critical role in making AI useful, by providing gold standard responses. The increasingly sophisticated behavior of AI models has made it easier for people with little-to-no computer training to interact with them \cite{daugherty2018human+}. 

In this paper, we present a humans-in-the-loop method for estimating the number of test items $N$, \emph{and responses per item} $K$, needed for \emph{reproducibly} estimating the performance difference between two AI models, while accounting for sampling variance across both items and responses per item, \emph{before more data are collected and models are retrained}.
This gives critical information about how to budget resources for building benchmark datasets. Our approach, which builds on methods from~\citet{wein-etal-2023-follow},
simulates the responses from a large pool of human raters and two ML models, rather than relying on methods that  
aggregate, and hence ignore, response variance.
Simulation enables us to generate enough response data to explore the significance boundary for NHST under various metrics, for $N$ test examples (items) with $K$ responses per item for each model and pool of human raters.
Our contributions are as follows:
\begin{itemize}
\item
While \citet{wein-etal-2023-follow} used their simulator to answer the questions of what are the best metrics and bootstrap configurations to use, they did not investigate the optimal trade-off in annotator budget between number of items $N$ and raters per item $K$. 

\item
The simulator of \citet{wein-etal-2023-follow} can only estimate $p$-values for NHST. We extend the simulator to also estimate the type-II error rate,
allowing for statistical power.

\item
We examine the trade-off between $N$ versus $K$ and report these results on seven real-world datasets;
by contrast, \citet{wein-etal-2023-follow} only investigated
\pv\ estimation
using a single dataset.
We show that these datasets, in their current size, lack enough responses for reproducibility of model performance. We further show that one can
boost the
reproducibility with fewer overall responses by collecting \emph{fewer items with more responses per item}. In fact,
our results in Section~\ref{sec:results} indicate that, for a fixed budget of $N \times K$ overall responses, apportioning the budget to as many as 100 responses per item can provide more reproducibility than with fewer responses per item.
\end{itemize}

\section{Related Work}
\label{sec:related-work}

Statistical testing
is critical to understanding state-of-the-art performance on a task or within a domain, in particular due to the \textbf{flawed nature of benchmarking practices} in machine learning evaluation \cite{ethayarajh2020utility,rajiai,rodriguez2021evaluation,hernandez2020ai}.
Existing statistical tests such as Student's t-test \cite{student1908probable} are based on strong assumptions,
such as that the datasets are normally distributed or have the same standard deviation,
which are not realistic, especially
when testing the system on new datasets \cite{sogaard-2013-estimating}. 
\citet{dietterich1998approximate} applied hypothesis testing to machine learning systems and \citet{dror2020statistical,deutsch2021statistical} provide a survey and guide to state-of-the-art techniques for statistical significance testing in AI systems.
\citet{longjohn2025statisticaluncertaintyquantificationaggregate} study the problem of aggregating across multiple tests. 
All of these studies apply to the case where each model yields a single response and a single correct label exists for each training example; therefore, the issue of response variance is ignored. 

More recently, \citet{gundersen_reproducibility_2020} exploited pseudo-random seeds to generate multiple model responses that could be used for improved statistical testing in the presence of a single correct label for each item.
\citet{10.1145/3186195} showed how to revise $p$-value calculation when ``gold'' annotations exist but are unknown and in their place multiple noisy ``bronze'' annotations are available, where the probability of a bronze annotation matching the gold is given.
In contrast, we consider settings where annotations are subjective and, hence, there is no single right answer but rather the ground truth is a distribution.

Our approach incorporates response variance from both ML models and human raters.
The nature of response variance of the former was studied in \citet{szymanski-gorman-2020-best}, claiming that
human rater response variance on individual items is most often due to measurable differences in perspective or ambiguity of the item, as opposed to noise.
Nuanced analysis of the nature of response variance in ML has been studied by
\citet{Artstein2008,plank-etal-2014-linguistically,peng-etal-2024-different,weerasooriya-etal-2023-subjective}.
See \citet{plank-2022-problem} for a survey. 

Although none of these methods have been widely adopted, beginning with \citet{dawid1979maximum}, researchers have recognized the importance of response variance, and have sought to characterize it. Most of these methods can be characterized as \textbf{tableau}-based, where items are visualized as rows and respondents as columns of an (often sparse) table \cite{passonneau-carpenter-2014-benefits}, and the models typically seek to jointly model both dimensions.  

\citet{lalor-etal-2016-building} apply item-response theory (IRT) to ML datasets. IRT is widely used in survey design and educational testing, two domains where, ironically, variance among respondents is widely reported, but variance among the items is not. (This makes sense for survey design, where each question addresses a different problem, but not in educational tests that contain multiple instances of the same problem, such as the Scholastic Aptitude Test (SAT).) And so they present a mirror image to the case of ML where, generally speaking, people tend to analyze variance along one dimension of the tableau, regardless of the domain, although \emph{which} dimension is used depends on the domain.

Related crowdsourcing studies have examined the trade-offs between cost and quality of annotation collection \cite{snow-etal-2008-cheap} or gave recommendations for which crowdsourcing platforms and protocols to use \cite{wang2013perspectives}. \citet{chau-etal-2020-understanding} explored the use of peer-review and self-review to resolve disagreement in annotations, and \citet{hovy-etal-2013-learning} developed an unsupervised model to identify which Mechanical Turk raters are reliable. Recent assessments of leaderboard practices have also led to models being able to indicate which items are most useful to annotate for evaluation purposes \cite{rodriguez2021evaluation}. \citet{welinder2010online} developed a system to select the most useful\slash informative labels to collect, which can lead to a reduction in annotation cost.

\citet{sheng2008getanotherlabel} focus on ML data curation and examines when one should obtain multiple, noisy training labels to improve model accuracy,
assuming there exists a single correct label for each example.
\citet{lin2014} claim that response variance is less important than item variance -- at least for training data --
and suggests collecting more items with a single response is more valuable than collecting multiple responses per item.

\citet{wein-etal-2023-follow} investigate \pv\ sensitivity of both metrics and test-set sampling methods in hypothesis testing,
which therefore can affect the power analysis.
While the latter did not turn out to be important in our study, metrics did.
Clearly, different metrics (e.g., mean absolute error vs Spearman rank-correlation) will produce different scores for the same matrix of responses, so it stands to reason that any comparison will have different \pvs\ for different metrics.
They model a metric as a function $\Gamma(M,G)$, where $M$ is a matrix of model predictions
which returns a score for $M$.
We assume $\Gamma$ is given here but focus on the best performing of these metrics in experiments.
\citet{homan2024how} initiates a study of the trade-off between number of items and responses using a toy simulator.
By contrast, we use real datasets to investigate these trade-offs and perform experiments that shed light on the mechanism for how response variance provides statistical significance. Recently, in a follow-up paper \cite{pandita2025forest} to this work, we modeled categorical datasets using a Bayesian approach and examined the optimization problem of allocating a fixed human annotation budget ($N \times K$). The current work provides the foundational evidence, for regression models, that increasing responses per item ($K$) is often more critical for significance than increasing the number of items ($N$).

The term \emph{multistage sampling} is commonly used in statistics when the data is subsampled at multiple levels of granularity, usually for stratification.
Bootstrap resampling has been applied in this setting~\cite{mashreghi16bootstrap} and so the sampling method we describe herein can be seen as an instance of these.
The Pigeonhole Bootstrap \cite{owen07pigeonhole} is quite different from our multistage bootstrapping
in that it resamples independently over rows and columns to form a Cartesian product rather than being nested.

It would be remiss not to mention other classes of techniques besides hypothesis testing that are commonly used for measuring statistical differences in model performance; see~\citet{riezler2021book} for a survey.
\emph{Likelihood ratios} provide an alternative form of significance testing and have been used for evaluating the impact of variability in data characteristics and hyperparameter settings on ML models~\cite{hagmann23inferentialreproducibility}.
Estimation statistics for reliability, most notably \emph{confidence intervals}, take variance into account to produce a range of values and are often used to assess a difference in model performance via non-overlap.
\emph{Circularity testing} based on general additive models has been proposed for evaluating the validity of ML models~\cite{riezler2021book}.

\section{Problem Statement}
\label{sec:problem}
We wish to apply null hypothesis significance testing (NHST) to compare the performance of
two ML models, $A$ and $B$, on a test set of $N$ items with $K$ responses per item 
and decide if one model is significantly better than the other.
We evaluate this with respect to human-annotated benchmark ``gold'' responses, $G$,
and according to a metric, $\Gamma$, which we assume is provided as a design hyperparameter.
For example, a common metric for evaluating regression models is the mean absolute error
(differences) between model predictions and gold annotations.

The null hypothesis assumes that the respective model output distributions are the same in relation to $G$.
Our goal is to determine whether the observations would be less than $5\%$ likely under the null hypothesis and, therefore, the null hypothesis can be rejected.
The $5\%$ level is what our calculated $p$-values are compared against to conclude significance.
Our motivation here is to determine whether a dataset---which we represent as $G^{N \times K}$, a matrix of $N$ items and $K$ responses---is large enough to provide replicable test results.
This can be applied either post-hoc, as a test of the reliability of results,
or at design time, before data is gathered and to help determine how best to allocate the usually limited amount of resources available for gathering human annotations.

A key innovation in this work
is to treat a data set $G^{N \times K}$ (as well as the responses from models $A$ and $B$) as a matrix of responses, instead of the pervasive simplifying assumption that $G$ is a vector, whose value for each item is an aggregation, such as the mean of several independent annotator (or model) responses.
The notation captures the further insight that the distribution of responses for each item in a dataset is different.

\section{Methods}
\label{sec:methods}

Our main contribution is a human-in-the-loop process that allows one to
(1) estimate the amount of data in terms of items, $N$, and responses per item, $K$, needed to detect, with high confidence,
a difference of performance according to metric $\Gamma$ of at least $\epsilon$; and
(2) compute \pvs\ for existing experimental data comparing the performance of two models against gold data.
Note that when the amount of experimental data is insufficient,
we can fit the data to a parameterized model and perform (1) to rerun the experiments with a sufficiently large dataset. It is precisely this use case that our experiments address.

Given an evaluation dataset $G$, arbitrary $N$ and $K$, $\epsilon > 0$ and metric $\Gamma$ the process has the following steps.
\begin{enumerate}
    \item Fit a two-stage probabilistic \emph{response model} model to $G$.
    \item Use that model via \emph{simulation} to determine p-values for $N$, $K$, $\epsilon$, and $\Gamma$.
\end{enumerate}
To fit a dataset to a response model, we create two histograms, one of all the individual responses over as a flat distribution and another of the average ratings of each item. We then find distribution families whose members visually match the distributions. Finally, we use the scipy package to find optimal parameters for the chosen model families fitting the dataset. See Section \ref{sec:data} and the appendix for more details. 

We then use a simulator to generate new gold responses as the same (fitted) distribution as $G$. 
We use the same given distribution to generate data for both $A$ and $G$, so that $A$ represents an ideal model for $G$. We add perturbation (governed by $\epsilon$) to this distribution to generate data for $B$. This ensures that model $A$ performs better than model $B$ with respect to $G$ under almost any metric,
and that ``ground-truth'' $p$-values should converge to zero as $\epsilon$, $N$, and/or $K$ increase.
The simulator then estimates $p$-values (or, in the case of power analysis $1-\beta$) based on a large number of repetitions $b$. Typically $b = 10000$, although power analysis requires two levels of repetitions: one to generate a distribution over effect sizes and one to estimate the \pv, given the effect size. We report the number of repetitions in the figures associated with each of our results (Figure \ref{fig:toxicity_wider_pval_for_NxK}).

The time complexity of computing \pvs\ in terms of the number of calls to the metric function $\Gamma$ is simply $b T(\Gamma)$, where $T$ measures the time complexity of $\Gamma$. For most of the choices of $\Gamma$ that we consider here, including MAE and Wins (see below), $T(\Gamma)$ is linear in the size of the matrix, hence the total complexity is $O(bNK)$.

\section{Experiments}
\label{sec:experiments}

\subsection{Data}
\label{sec:data}
Unfortunately, precious few public datasets have both a large number of items and disaggregated responses.
We apply the metrics and \pv\ estimators to the following datasets, all of which are secondary to us. We essentially ignore the content of each item in each dataset and use only the human responses associated with each item. Even though these responses were generated by humans--and we believe modeling human annotators is a promising direction to explore---to simplify our analysis and minimize risk we ignore any information about those humans and treat the responses for each item as, effectively, an anonymous sample.

In the experiments, we use the data to fit parameterized models. This allows us to study the performance (counterfactually) of the metrics under different values of $N$, $K$ than the ones inherent to datasets, and for different values of $\epsilon$ due to different (hypothetical) models. We need to rely on counterfactuals and hypotheticals, even though we have real data, because no extant dataset has enough responses for large enough $N$ and $K$ or models with specific $\epsilon$ for us to run our experiments, and collecting that data would be prohibitively expensive. In fact, the \emph{motivation behind this research} is precisely the problem that we need to choose reasonable values for $N$ and $K$ \emph{before} we collect data, because no one has the budget to collect data for arbitrary values of $N$ or $K$.

The \textbf{MultiDomain Agreement} \cite{leonardelli2021agreeing} dataset contains tweets about Black Lives Matter, the US 2020 presidential election, and COVID-19, annotated for offensiveness. The test set has 3057 items annotated by 5 raters each.
We fit the means and standard deviations of the item responses to \emph{truncated} normal distributions with ($\mu=-0.5, \sigma=1)$ and $(\mu=-0.3923, \sigma=0.8502)$, respectively.
Instructions for directly obtaining the dataset from the author are available at \url{https://github.com/dhfbk/annotators-agreement-dataset}.

The \textbf{Stanford Toxicity} dataset~\cite{kumar2021designing} was also used in \citet{wein-etal-2023-follow}.
It contains 107,620 items annotated by 5 raters each with ratings on a 5-point Likert scale: not/slightly/moderately/very/extremely toxic.
We use the same distributions as they do, namely, a folded normal with $(\mu=0.19, \sigma=0.11)$ for the means and a triangular distribution with $(a=-0.05, b=0.21, c=0.45)$ for the standard deviations. The data is available at \url{https://data.esrg.stanford.edu/study/toxicity-perspectives}. It is encrypted, but the website gives instructions for how to decrypt it. There is no published license.

\subsection{Fitting the Simulator to Real Data}
\label{appendix:fitting-simulator}
The simulator allows us to generate many test sets to extrapolate patterns beyond one domain or system. By holding the item distributions for $A, B$, and $G$ fixed, we can draw from them repeatedly to generate test sets similar to a real dataset but with arbitrarily large values of $N$ and $K$, which would be infeasible with actual human annotations.

Like~\citet{wein-etal-2023-follow}, for each set of responses (from models $A$ or $B$, or $G$), we sample from multistage parameterized models to simulate multiple samples for fixed $N$ and $K$ from a data source.
This multistage process uses two probabilistic models, where for each item $i$ the second stage model generates responses for the item $P(i)$, while the first stage model generates for $i$ parameters unique to $i$ for the second stage model to generate each response (i.e., $P(j|i)$ for response $j$ to item $i$).
In contrast to \citet{wein-etal-2023-follow}, we choose the parameterized models to fit real datasets. Each dataset has enough responses \emph{over all items} for us visualize the \emph{a priori} distribution (i.e., $P(j)$ for item $j$, without regard to the item $i$ it is associated with), say, as a histogram and use that to make informed choices about what families of parameterized distribution might fit the data. However, none of these datasets has enough responses \emph{per item} for us to conclude anything about the shape of the \emph{prior distribution} of responses for any item (we are not aware of any dataset that has both enough gold responses per item to visualize responses). And so for the second state model, we apply the principle of maximum entropy and assume the per-item distribution of responses is a \emph{generalized normal distribution} $\normdist(\mu_i, \sigma_i)$. With more data per item, we could easily swap in a different family of distributions if we observed meaningful patterns in per-item responses.

\begin{figure}
\centering
\includegraphics[width=\linewidth]{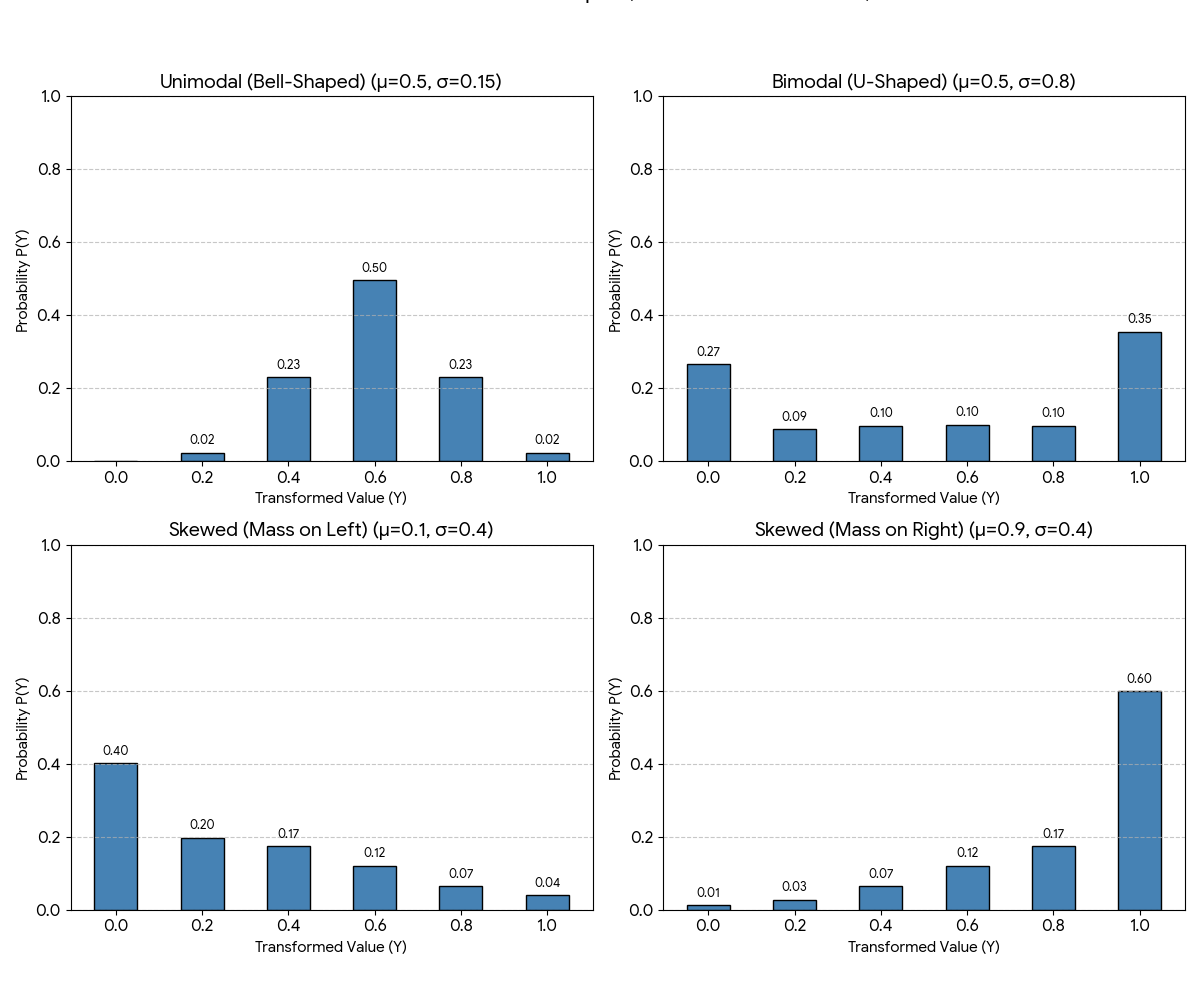}
\caption{
Shapes possible with censored normal.
}
\label{fig:censored_normal_shapes}
\hspace{.3cm}
\end{figure}

However, because we do have enough responses \emph{over all items}, we do choose for the first stage specific distributions for each dataset that, paired with the second stage described above, fit the data. We used the \emph{censored} normal distribution for $\normdist$,
which assumes a latent continuous distribution that is not observed exactly
but measured to within intervals, including left and right intervals which
\emph{pool} (not truncate) the smallest and largest values, respectively.
This provides support for head and/or tail bias;
Figure~\ref{fig:censored_normal_shapes} illustrates a variety shapes that this distribution can capture.
For example, items in the Stanford Toxicity dataset
(see Section \ref{sec:data}) rated at either extreme (either ``not toxic'' or ``extremely toxic'')
tend to have more agreement among raters.

\begin{figure}[htb]
\centering
\begin{subfigure}[]{\linewidth}
\centering
\includegraphics[width=\linewidth]{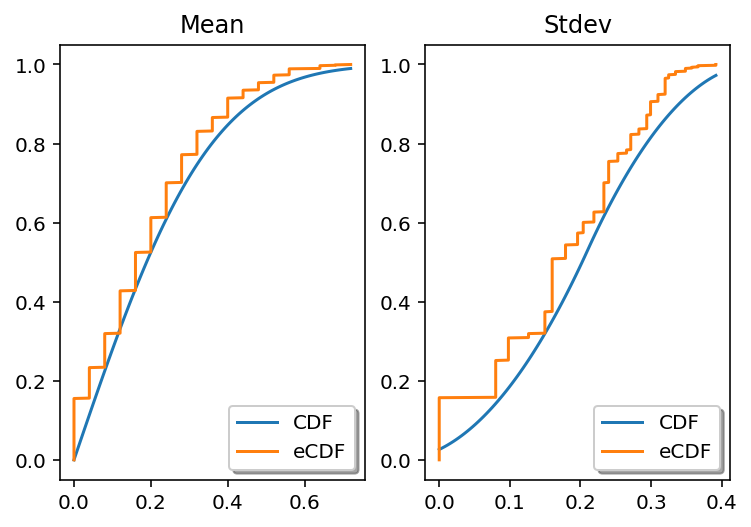}
\caption{Stanford Toxicity}
\label{fig:toxicity_cdf}
\end{subfigure}
\begin{subfigure}[]{\linewidth}
\centering
\includegraphics[width=\linewidth]{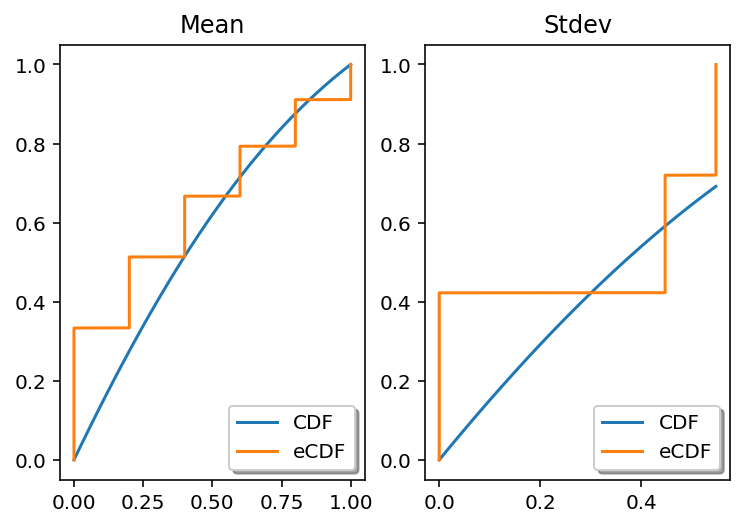}
\caption{MultiDomain Agreement}
\label{fig:mdagreement_cdf}
\end{subfigure}
\caption{
Empirical CDFs of item-level response means and standard deviations in
(a) the Stanford Toxicity dataset vs clipped, folded normal CDF with
$\langle \mu=0.19, \sigma=0.11 \rangle$
and clipped triangular distribution CDF with
$\langle a=-0.05, b=0.21, c=0.45 \rangle$, respectively; and
(b) the MultiDomain-Agreement dataset vs truncated normal CDF with
$\langle \mu=-0.5, \sigma=1 \rangle$
and truncated normal CDF with
$\langle \mu=-0.3923, \sigma=0.8502 \rangle$, respectively.
}
\label{fig:goodness_of_fit}
\end{figure}

We use distributions fitted to each dataset from distribution families tailored to each dataset.
Note that we chose the folded and truncated normal and triangular distributions for these datasets \emph{based on visually matching histograms of the responses of each dataset}, as described in Section \ref{sec:methods}. We can use any family of distributions we like, i.e., \emph{they need not be any flavor of normal distribution}, as long as there are algorithmically feasible ways of fitting them to the data. 
Figure~\ref{fig:goodness_of_fit} illustrates goodness-of-fit for simulations of two datasets used in this paper. Details about computing \pvs\ and results for additional datasets can be found in Appendix \ref{sec:est_p_values} and \ref{appendix:additional_results}.

\subsection{Results}
\label{sec:results}
We mainly used the following metrics in experiments:
\begin{itemize}
\item \textit{Mean absolute error difference} (MAE).
The distances (errors) from the per-item mean gold response to the model response averaged over the items:
$\Gamma_{\rm MAE}(A,B,G) =
\frac{1}{N} \sum_i^N \bigl(
\left| \frac{1}{K} \sum_j^K B_{ij} - \frac{1}{K} \sum_j^K G_{ij} \right| - \allowbreak
\left| \frac{1}{K} \sum_j^K A_{ij} - \frac{1}{K} \sum_j^K G_{ij} \right|
\bigr)$
\item \textit{Item-wise wins} (Wins).
The fraction of items in the test set for which the absolute error of A is smaller than B: 
$
\Gamma_{\rm Wins}(A,B, G) =  \sum_{i = 1}^{N} \mathbf{1}_{<} ( |\overline{A_i} -\overline{G_i}|, |\overline{B_i} - \overline{G_i}| )/N
$
\item \textit{Mean EMD difference} (MEMD).
The Earth mover's distance for each item between the system and the gold standard responses, and then take the mean of those item-wise EMDs:
$\Gamma_{\rm MEMD}(A,B,G) = \sum_{i = 1}^N \left( {\rm EMD}(B_i, G_i) - {\rm EMD}(A_i, G_i)\right)/N$
\end{itemize}


\hide{
All four of these datasets differ from the synthetic data in that \mae\ 
outperforms \wins.   
Also, none of these test sets appears to be large enough to distinguish between two models with $\Delta_{\rm WINS} < 0.01$, which is a relatively large performance gap to see near the top of a leaderboard.
}

\hide{
The distribution of the item-wise means (\Cref{fig:fitting_means}, left) seems to follow a folded normal distribution that has been \emph{clamped} to the range $[0,.8]$ (i.e., values falling outside that range are assigned to the nearest value in the range, namely $0$ or $1$). The standard deviations (\Cref{fig:fitting_stds}, right) seem to follow a triangular distribution clamped to the range $[0, \infty]$ (i.e., only values less than $0$ are reassigned). 
Table~\ref{tab:toxicity_pvalues} shows an abbreviated version of these results.
As with the LeWiDi datasets,
\mae\ is more sensitive than  \wins.
We also report accuracy for this metric, as that is the metric used in~\cite{kumar2021designing}.
}

\begin{figure*}[htb]
\centering 
\begin{subfigure}[]{\linewidth}
\centering
\includegraphics[width=5in]{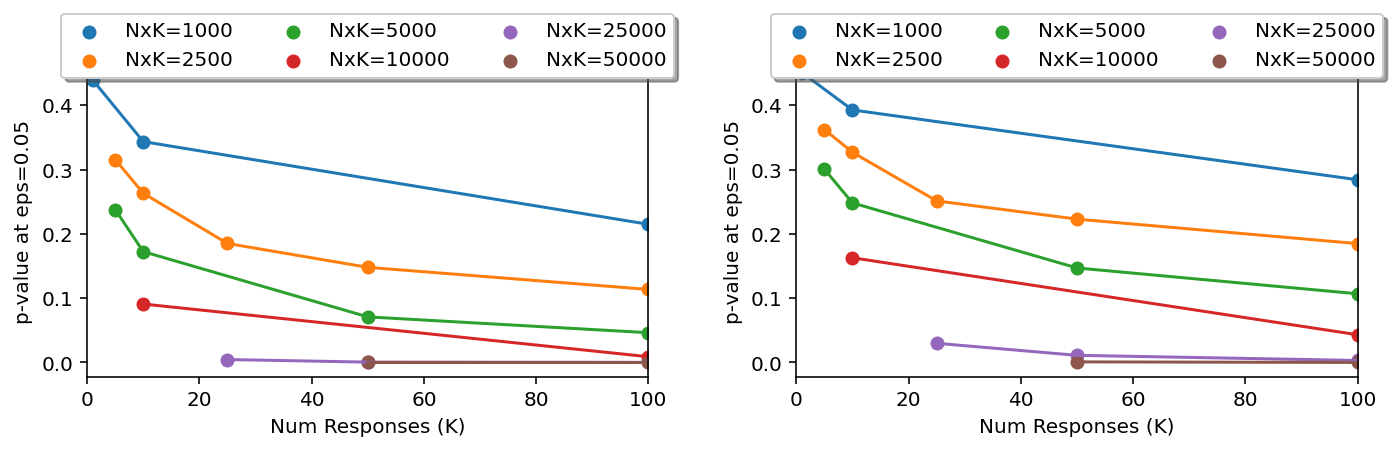}
\caption{Toxicity ($\epsilon=0.05$): MultiStage (left) vs ``flat'' bootstrap (right)}
\label{fig:toxicity_pval_for_NxK}
\end{subfigure}
\hspace{.3cm}
~\\
\begin{subfigure}[]{\linewidth}
\centering
\includegraphics[width=5in]{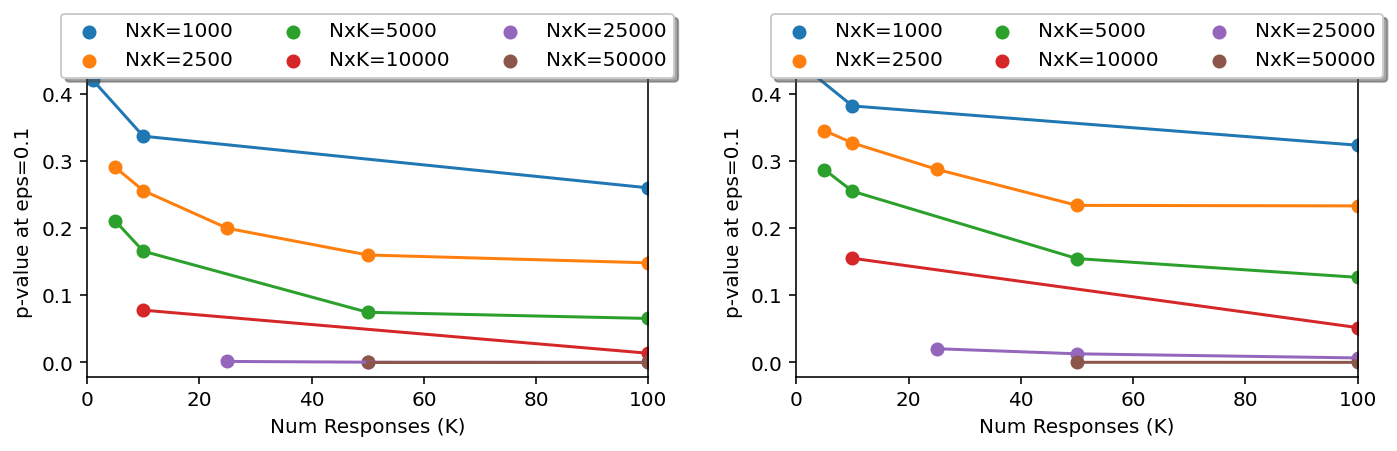}
\caption{MultiDomain ($\epsilon=0.1$): MultiStage (left) vs ``flat'' bootstrap (right)}
\label{fig:mdagreement_pval_for_NxK}
\end{subfigure}
\caption{
\pv\ vs $K$ with \mae\ at various $N \times K$. Each data point is the estimated from $10,000$ samples. 
}
\label{fig:pval_for_NxK}
\end{figure*}

\begin{figure}[htb]
\centering
\includegraphics[width=\linewidth]{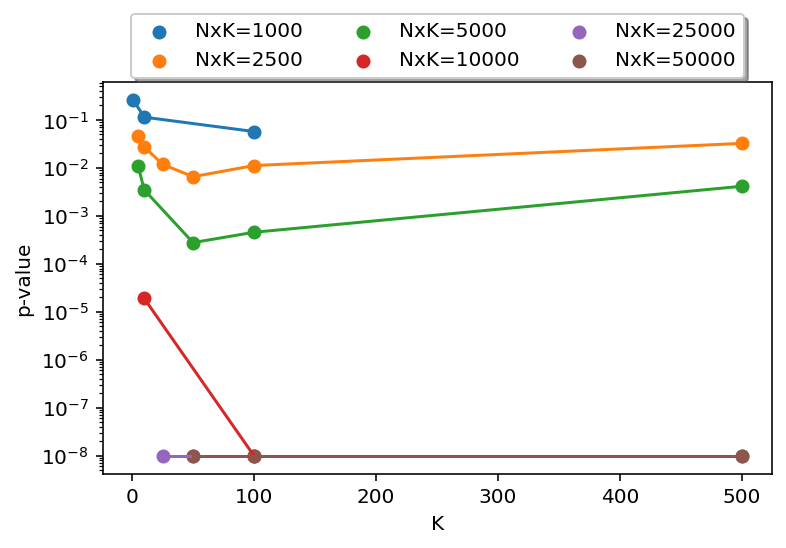}
\caption{
\pv\ vs $K$ with \mae\ at various $N \times K$ for
Toxicity at log-scale on the y-axis. Each data point is the estimated from 10000 samples. 
}
\label{fig:toxicity_wider_pval_for_NxK}
\hspace{.3cm}
\end{figure}

\begin{figure}[htb]
\centering
\begin{subfigure}[]{0.85\linewidth}
\centering
\includegraphics[width=\linewidth]{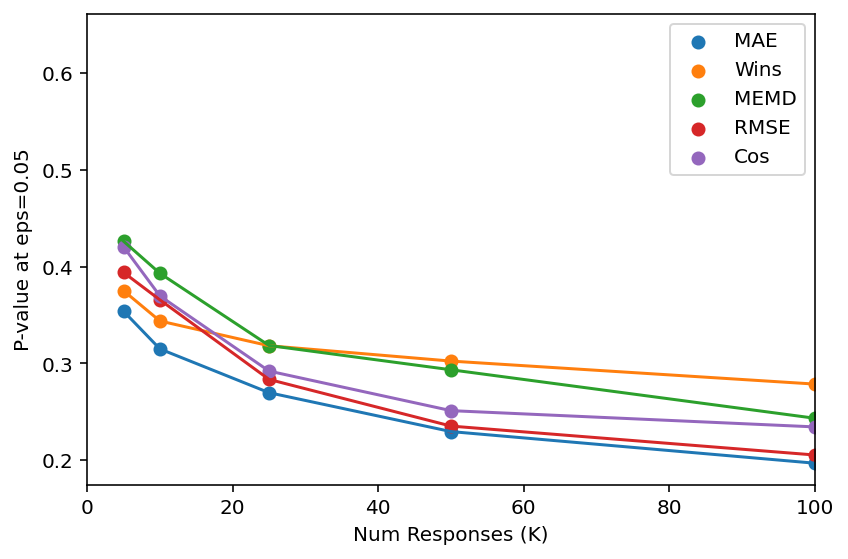}
\caption{Toxicity ($\epsilon=0.05$)}
\label{fig:toxicity_various_metrics}
\end{subfigure}
\hspace{.3cm}
\begin{subfigure}[]{0.85\linewidth}
\centering
\includegraphics[width=\linewidth]{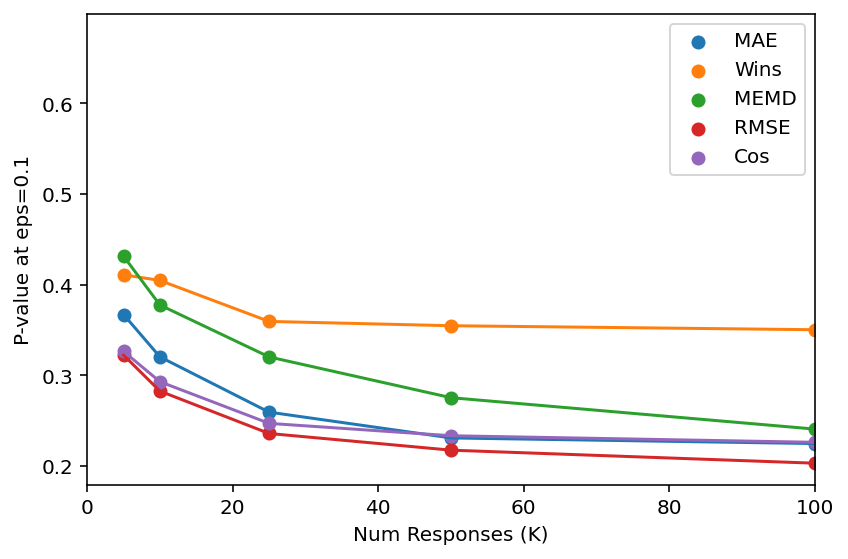}
\caption{MultiDomain ($\epsilon=0.1$)}
\label{fig:mdagreement_various_metrics}
\end{subfigure}
\caption{
\pv\ vs $K$ with a fixed budget $N \times K = 2500$ for various metrics. Each data point is estimated from $10,000$ samples. 
}
\label{fig:various_metrics}
\end{figure}

We utilize the Variance Estimation Toolkit (VET)\footnote{\url{https://github.com/google-research/vet}} to run our experiments. We used the Python libraries NumPy, Pandas, and SciPy, versions 2.2.3, 2.2.1, and 1.13.1, respectively. Our experiments took various times to run, with the longest experiments (producing any of the points in our figures) running approximately nine hours.

Figure~\ref{fig:pval_for_NxK} demonstrates that trading off items for responses is beneficial at a wide range of $(N \times K)$ values, with $p$-value decreasing as $K$ increases.
(The benefit of increasing $K$ is strikingly more apparent when viewing $p$-values vs $K$ with a fixed $N$, but we omit these graphs for brevity.)
Here \mae\ was used with distortion $\epsilon=0.05$ for Toxicity and $\epsilon=0.1$ for MultiDomain, but similar trends were observed using other metrics, amounts of distortion, as well as different datasets.
The graphs on the left are based on using the multistage bootstrap whereas those on the right
use the baseline ``flat'' bootstrap over only the items, after the per-item responses have been aggregated.
Note that the multistage bootstrap \pvs\ are smaller, hence closer to the ground-truth values of zero,
as it makes better use of response variance.
There is indeed a point where trading $N$ for $K$ is beneficial for statistical significance: in this case, the curves hit an inflection point before $K=500$; see Figure~\ref{fig:toxicity_wider_pval_for_NxK}.

Figure~\ref{fig:various_metrics} graphs \pv\ as a function of number of responses at $\epsilon=0.1$, where the number of items varies such that $N \times K = 2500$, and demonstrates a similar trend across five different metrics.

\subsubsection*{Power Analysis}
\begin{figure}[htb]
\centering
\begin{subfigure}[]{0.85\linewidth}
\centering
\includegraphics[width=\linewidth]{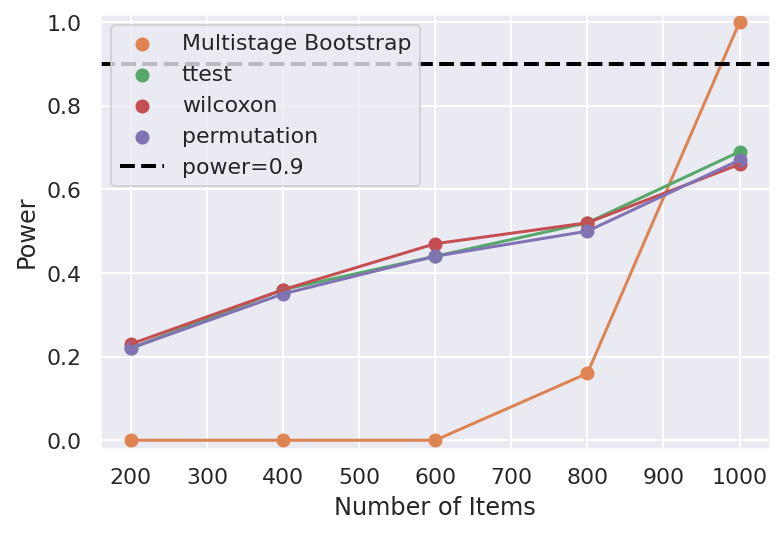}
\caption{Varying $N$ with $K=5$}
\end{subfigure}
\begin{subfigure}[]{0.85\linewidth}
\centering
\includegraphics[width=\linewidth]{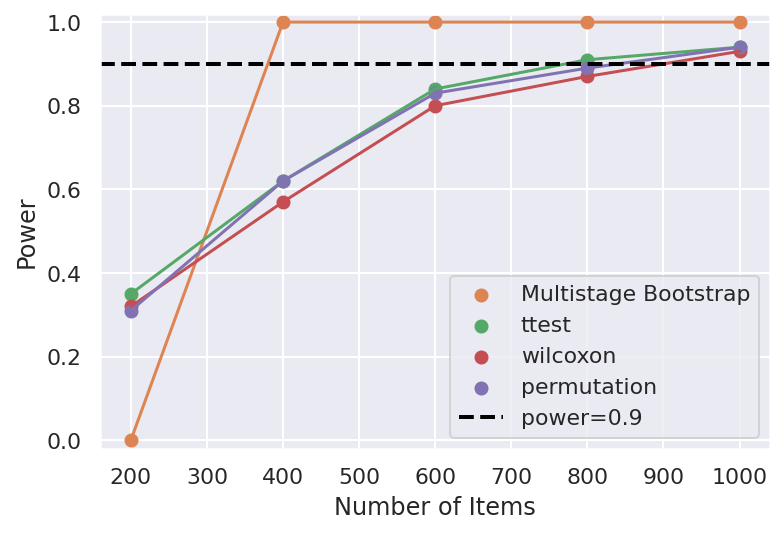}
\caption{Varying $N$ with $K=10$}
\end{subfigure}
\begin{subfigure}[]{0.85\linewidth}
\centering
\includegraphics[width=\linewidth]{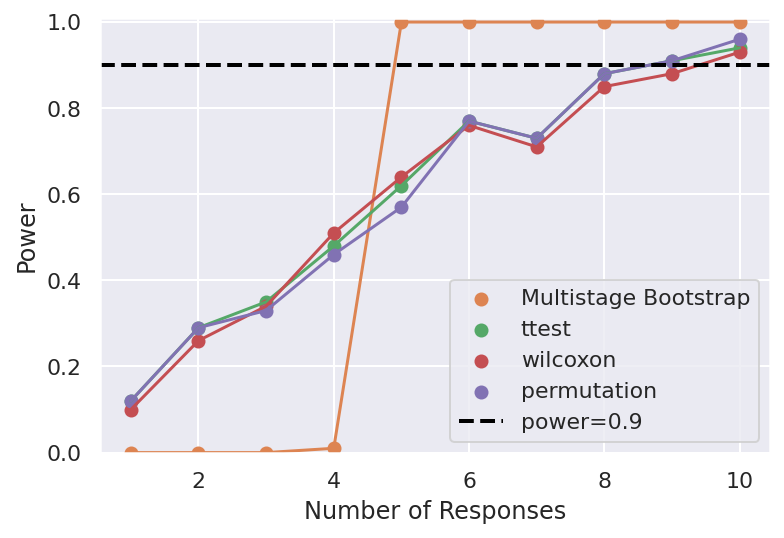}
\caption{Varying $K$ with $N=1000$}
\end{subfigure}
\caption{Power Analysis of Toxicity data ($\epsilon=0.1)$. Each data point is the estimated from 1000 outer-level samples, each consisting of 10000 inner level samples.}
\label{fig:toxicity_power_vs_sample_size}
\end{figure}

\hide{
\begin{figure}[htb]
\centering
\begin{subfigure}[]{0.85\linewidth}
\centering
\includegraphics[width=\linewidth]{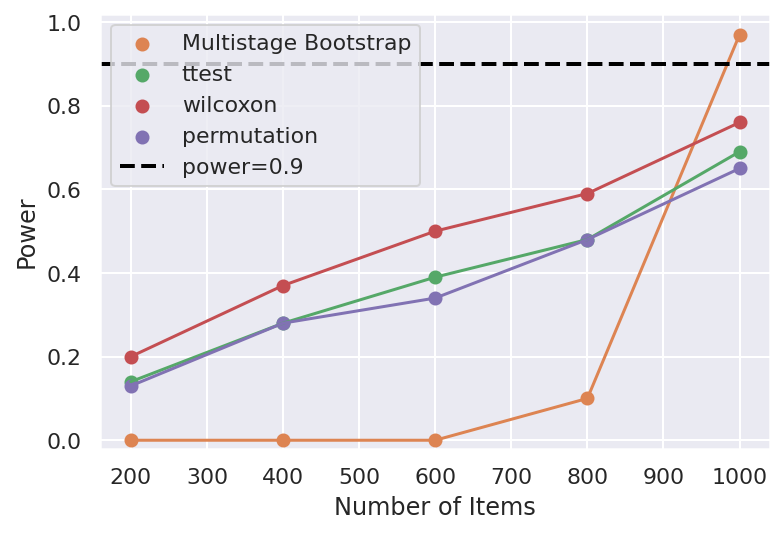}
\caption{Varying $N$ with $K=15$}
\end{subfigure}
\begin{subfigure}[]{0.85\linewidth}
\centering
\includegraphics[width=\linewidth]{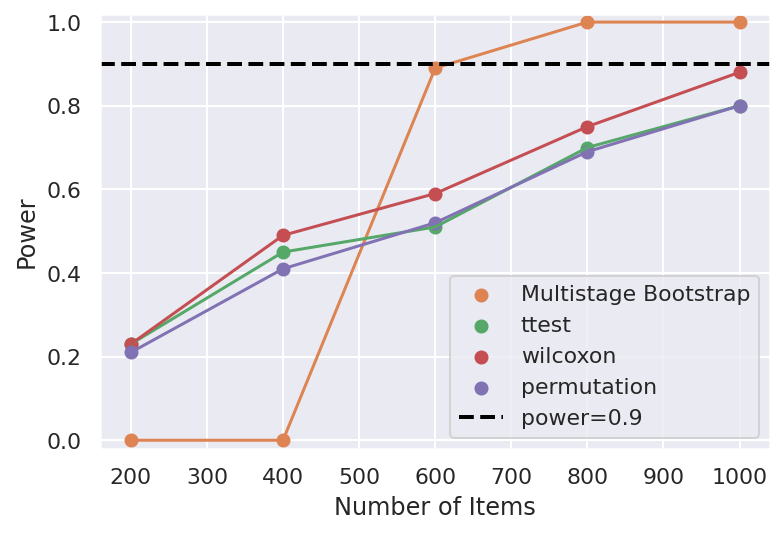}
\caption{Varying $N$ with $K=20$}
\end{subfigure}
\begin{subfigure}[]{0.85\linewidth}
\centering
\includegraphics[width=\linewidth]{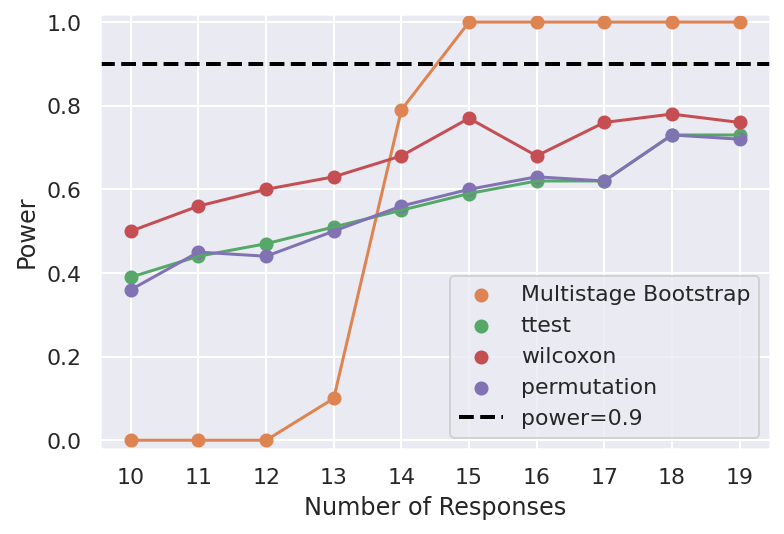}
\caption{Varying $K$ with $N=1000$}
\end{subfigure}
\caption{Power Analysis of MultiDomain data ($\epsilon=0.1)$. Each data point is the estimated from 1000 outer-level samples, each consisting of 10000 inner level samples.}
\label{fig:md_power_vs_sample_size}
\end{figure}
}

Figure~\ref{fig:toxicity_power_vs_sample_size}
demonstrate greater statistical power for Multistage Bootstrap as sample size
with respect to either number of items or responses
increases, achieving a power of 90\% (i.e., probability of not rejecting the null hypothesis when it's false) before baseline hypothesis tests.
As usual, we use $\alpha=0.05$ as the significance level for power calculation, i.e., the data is inconsistent with the null hypothesis at least 95\% of the time.
While the power of all these tests benefit from having more responses, the rate of improvement is markedly more rapid for Multistage Bootstrap.

For the baseline (paired) hypothesis tests, the mean response of each item was pre-computed for Model A, Model B and for ``gold'' G, resulting in $\bar{a}_i, \bar{b}_i, \bar{g}_i$, respectively, for each item $i$. The baseline tests then consider the null hypothesis that the distributions across the items of $|\bar{a}_i - \bar{g}_i|$ and $|\bar{b}_i - \bar{g}_i|$ are the same in the case of the permutation test, or have the same center in the case of Welch's t-test and the Wilcoxon signed-rank test.
In contrast, Multistage Bootstrap resamples both the set of items and, for each item, the set of responses at each iteration, hence more effectively taking into account the disaggregated distribution of responses.

\section{Discussion}
\label{sec:discussion}
Our results indicate that the number of raters and items have a notable impact on \pv\ estimation, to different degrees depending on the metric.
\wins\ provides a discrete decision for each \textit{item}, counting those decisions (i.e.``wins'') across the test set and normalizing by the number of items.
\wins\ is also presented as a meta-metric of sorts: it can use any item-level metric, with absolute error being used here, and requires both models' predictions as well as input to directly compare their predictions at the item level.

In general, increasing $N$ (number of test set items) increases the statistical power of any measurement by simply providing more scores to base the final metric score on. The more scores there are, the more stable the variance across simulation runs will be, and the lower the \pv. All examined metrics respond well to increasing $N$.

Increasing $K$ (number of responses per item) increases the statistical power of each \textit{item level aggregate}.  As $K$ increases, the lower the variance of an individual item's aggregate will be across simulation runs, thereby lowering the \pv. 
All tested metrics also respond well to increasing $K$.

The difference between the metrics lies in the way the item-level scores are used.  For Wins, which responds better to increasing $N$, the $A$'s and $B$'s item-level scores are directly compared.  In each run, these item-level scores will vary, but in many cases that variance won't change the pairwise comparison. For example, if $A_i$'s metric score is 0.10 and $B_i$'s is 0.12 on the first simulation, a win is recorded for $A$. In the next simulation, if the scores are 0.11 and 0.12, respectively, this score change does not change the Win, as $A_i$'s score is still lower. This indicates the item-level variance in the discrete win decision is far lower than the score variance - so adding more responses is less likely to further reduce the variance than adding items.

By contrast, for $\Gamma_{\rm MAE}$ and $\Gamma_{\rm MEMD}$, any changes in item-level metric scores do impact the variance, both at the item and test-set level. Since the item-level scores come from the response distribution, adding more responses stabilizes the simulated distributions under repeated test set generation, reducing the metric variance across simulations and lowering the \pv. 

The implications of these results are that the item\slash response trade-off should be handled differently depending on the metric itself, and the demands on the number of raters and items are high for all metrics in order to provide statistical guarantees. However, our results suggest that shifting the budget to account for as many as 100 raters per item could improve the sensitivity of experiment data to effect sizes.

\hide{
As to why $\Gamma_{\rm MAE}$ was more sensitive than $\Gamma_{\rm WINS}$ on the real-world data but the reverse was true on the synthetic data. We suspect that this is due to the discrete nature of these responses versus the continuous nature of the synthetic responses, however more research is needed to confirm this. 
}

The datasets we explored here have simple output domains. What about generative models whose outputs may be highly complex? In this case, the ratings tend to be specific for each model, provided by a human or, increasingly, another generative model. And the ratings tend to be simple. In this case, our methods can be adapted. Essentially, we drop the gold dataset and only model the distribution of ratings received for each models \emph{not the actual model responses}. This also requires a different comparision metric that does not require a gold dataset and instead of model responses takes as input the ratings each model received for its outputs. However, the basic process is the same.

\section{Conclusion}

In this work, we experimented with simulated data in order to examine the trade-off between the number of items and the number of responses per item necessary to compare two models against human judgments with statistical significance ($p<0.05$). As expected, we see that when two models are more similar in performance, a greater number of annotations is required to achieve significance on their comparison. Further, the metric itself affects the utility of an increase in either items or responses.

These results suggest that current evaluation practices are not sufficient to confidently assess two models' performance against gold judgments, as using 25,000-50,000 annotations in a test set is rarely seen. Even when using 1000 items, at least 25 raters are needed for models to achieve significance with MAE.

Additionally, we found that the trade-off between the number of items and the number of responses per item depended on the metric. For two of our tested metrics, MAE and mean EMD, adding more responses than items is a more optimal division to achieve lower \pvs. For the Wins metric, the opposite is true: more items and fewer responses per item lead to lower \pvs. Still, in all cases for all metrics, increasing the total number of responses consistently lowers \pvs, and thereby increases the sensitivity of the evaluation instrument. For real-world data, we actually found MAE to be more sensitive than Wins.

\hide{
The main takeaway from all of this is that different metrics are more sensitive to \pvs\ than others, but which metric is more sensitive is situational. This suggests that sensitivity to significance is something to consider when choosing which metrics to use to evaluate ML performance, in addition to more commonly-considered properties such as interpretability or feasibility. Our results are a proof-of-concept that the framework of \cite{wein-etal-2023-follow} can be used to evaluate which metric is most sensitive in a specific real-world setting.
}

\section*{Limitations}
\label{sec:limitations}
The effectiveness of \citet{wein-etal-2023-follow}'s simulator depends on how well the probabilistic models capture realistic distributions of responses over items. Although we used rigorous methods to fit the parameters of these distributions to our datasets, our choice of distribution family to use for each dataset was based on visual inspection of the data. Given more datasets with disaggregated responses, we hope in future work to develop rigorous methods for model selection. However, the dearth of such publicly-available datasets impedes progress in this direction. One key limitation future work will address is that we treat the responses as independent from item-to-item, when in reality responses usually depend on which human annotator or instance of a model produced the response. Hypothesis testing such as that described here is not a comprehensive measure of data quality; it only estimates the likelihood of sampling error. It does not account for sampling bias, leading to data that is not representative of the sampling distribution.

The simulator is only intended to capture the complexity of the annotations.
It is not intended to capture the complexity of real model predictions but rather to compare
a near-perfect model, $A$, against a version, $B$, that has been perturbed by a controlled amount
via a variance parameter.
In practice, this functions as an approximate bound on the model response variance.

Otherwise, we have taken precautions to avoid common ``$p$-hacking'' pitfalls, such as
that the null hypothesis and significance threshold $\alpha$ are independent of the dataset.
We attempt to avoid \emph{optional stopping} by performing power analysis.

While the distribution of responses depends on each item, we do not assume a fixed correspondence between annotations and raters.
This assumption is valid, for example, with a large rating pool where each rater annotates at most one item. 
Therefore, there is no meaningful ordering of the responses within each item.
For convenience, we use the term ``matrix'' for what is really a sequence of multisets.
Modeling the dependence of annotations from the same raters across multiple items
is something we chose to ignore in this paper
so as not to distract from its main focus on the impact of response variance on hypothesis testing.

\section*{Ethical considerations} The paper focuses on a method to ensure that enough data is collected during testing to ensure that large enough observed differences between the performance of two models on the data are significant. While such analysis can ensure that experiment results are meaningful and replicable, p-values have a tendency to be used more than they are understood. It is important to understand what p-values guarantee and what the limitations of our, or any other particular NHST framework, are. Misinterpreting the analysis can lead to dishonest or misleading claims about the reliability of the data for testing.

\bibliography{custom}


\appendix

\section{Using the simulator to estimate $p$-values}
\label{sec:est_p_values}

\subsection{Simulator}
\label{sec:simulator}

We use a simulator
to generate ``gold'' annotations and model predictions by modeling the responses for each item as a random variable.
The purpose of this is to be able to control how similar, or different, predictions from models $A$ and $B$ are to $G$ as well as to each other.
By using the same given distribution to generate data for both $A$ and $G$,
and by adding perturbation (governed by parameter $\epsilon$) to the given distribution to generate data for $B$,
we can ensure that model $A$ performs better than model $B$ with respect to $G$ under almost any metric,
and that ``ground-truth'' $p$-values should converge to zero as $\epsilon$, $N$, and/or $K$ increase.

The simulator takes input parameters $N$ and $K$, along with \emph{perturbation parameter} $\epsilon$.
In the first stage, it randomly chooses
hyperparameters $\mparams_1, \ldots, \mparams_N \sim \itemdist$,
each corresponding to an item $\mparams_i$,
from a fixed distribution that serve as model parameters for the second stage.
In the second stage, for each item $i$
we sample $K$ responses from a second distribution $\respdist(\mparams_i)$.
We do this for each of the datasets, respectively representing responses from gold annotations, $G^{N\times K}$, and two models, $A^{N\times K}$ and $B^{N\times K}$.
\hide{
For each of $N$ items in the test set, the simulator randomly draws a mean and standard deviation $\{(\mu_i, \sigma_i)\ i \in [1,N]\}$, where $\mu_i \sim \mathcal{U}[0,1]$, $\sigma_i \sim \mathcal{U}[0,.3]$, $\mathcal{U}$ is the uniform distribution and $\sim$ indicates a random draw from a distribution. It then draws from the resulting normal distribution $\mathcal{N}(\mu_i, \sigma_i)$ $K$ times to produce the gold standard set of responses $G_i$ (clipping values outside of $[0,1]$). This per-item draw of $K$ responses is repeated to produce the model predictions $A_i$. This models the idealized situation in which model $A$ is a perfect representation of the gold standard, since \textit{it is drawn from the same distribution}.
A third set of item responses for model $B$ is then drawn by injecting random noise into the base distribution of each item.
For a given perturbation level $\epsilon$, which we choose, a noise parameter is randomly drawn for each item $\epsilon_i \sim \mathcal{U}[-\epsilon, +\epsilon]$, and then $K$ responses for each $B_i$ are drawn from $\mathcal{N}(\mu_i + \epsilon_i, \sigma_i)$.
}
The specific distributions that were used in our experiments were modeled from real datasets;
for details see Appendix~\ref{appendix:fitting-simulator}.

These choices operationalize a solution to the paradox that one must have data in $G$, $A$, and $B$ to know if it has enough statistical power. Instead, we simulate a set of gold items and responses ($G$) and then simulate an ideal model ($A$) -- ideal because it draws its simulated responses from the same distribution as the gold -- and then explore how such an ideal system would compare in significance to another model ($B$) whose response distributions differ from gold by an amount ($\epsilon$) we experimentally control. This gives us \emph{a-priori} control over the hypothesis test, because we know which model is better through a controllable parameter.

For any given selection of $N$ and $K$, we have response matrices $G^{N \times K}$ and $A^{N \times K}$ and, for each $\epsilon$, a matrix $B^{N \times K,\epsilon}$. We then seek to compare $A$ and $B$ to each other to determine which is better; the answer should almost always be $A$ unless $\epsilon=0$.  When evaluating AI models, the comparison of $A$ and $B$ involves differencing each of their item responses to those of $G$ using a suitable metric,
which is then aggregated across the items.
We
compare the performance between $A$ and $B$ via $\Gamma(A, B, G)$.

\hide{
The simulator allows us to generate many test sets to extrapolate patterns beyond one domain or system.
With this variance, we can construct a null hypothesis set and measure how likely it is that an observed difference between the two metric scores could have occurred by chance.
We perform experiments on different simulated datasets for various combinations of $N$ and $K$. 
}

\hide{
\citeauthor{wein-etal-2023-follow} analyze several metrics,
suggesting a few that give the lowest \pvs. We chose three of the best-performing metrics: mean absolute error (MAE), item-wise wins (Wins), and Mean EMD (MEMD):
\begin{itemize}
\item \textit{Mean absolute error difference} (MAE).
The absolute value of the distance (error) from the mean gold responses per item to the mean system responses, and then take the mean of that item-wise error:
$$\Gamma_{\rm MAE}(A,B,G) = \overline{\{|\overline{B_i} - \overline{G_i}|} - \overline{|\overline{A_i} - \overline{G_i}|\}}_{1\leq i \leq N}$$
\item \textit{Item-wise wins} (Wins).
The fraction of items in the test set for which the absolute error of A is smaller than B: 
$$
\Gamma_{\rm Wins}(A,B, G) =  \sum_{i = 1}^{N} \mathbf{1}_{<} ( |\overline{A_i} -\overline{G_i}|, |\overline{B_i} - \overline{G_i}| )/N
$$
\item \textit{Mean EMD difference} (MEMD).
The Earth mover's distance for each item between the system and the gold standard responses, and then take the mean of those item-wise EMDs:
$$\Gamma_{\rm MEMD}(A,B,G) = \sum_{i = 1}^N \left( {\rm EMD}(B_i, G_i) - {\rm EMD}(A_i, G_i)\right)/N$$
\end{itemize}
}

\subsection{Estimating $p$-values}

Given $N$, $K$, and $\epsilon$, \pvs\ are estimated by drawing $b$ (bootstrap) resamples
$S_{alt} = \langle G^{N \times K}_1, A^{N \times K}_1, B^{N \times K}_{1,\epsilon}\rangle,$ $\ldots,$ $\langle G^{N \times K}_{b}, A^{N \times K}_{b}, B^{N \times K}_{b,\epsilon}\rangle$
for the alternative hypothesis
according to the process described in Section \ref{sec:simulator}.
Since the null hypothesis makes the assumption that the distributions of $A$ and $B$ are the same with respect to $G$,
we construct $S_{null}$ by pooling the items from $A^{N \times K}$ and $B^{N \times K}$
and then independently sampling from this pool.
When sampling responses from $A$,
for each item $i$, we sample each response by
sampling from $\respdist(\mparams_i)$, where $\mparams_i = (\mu_i, \sigma_i)$.
Sampling responses from $B$ is similar but we first
choose $\delta_i \sim Unif(-\epsilon, \epsilon)$ and then
sample from $\respdist(\mparams_i)$, where $\mparams_i = (\mu_i + \delta_i, \sigma_i)$.

Next, we estimate the \emph{expected $p$-value under the alternative hypothesis} as the average one-sided \pv\ over all samples in $S_{alt}$, computed by counting for each $s_{alt} = \langle G^{N \times K}_{alt}, A^{N \times K}_{alt}, B^{N \times K}_{alt,\epsilon}\rangle \in S_{alt}$ the fraction of samples $s_{null} \in S_{null}$ where $\Gamma(s_{null})$ is at least as extreme as $\Gamma(s_{alt})$. Here ``at least as extreme'' is determined by computing $\Gamma_{alt}$ (respectively, $\Gamma_{null}$), the median of $\Gamma$ over $S_{alt}$ (respectively, $S_{null}$). If $\Gamma_{alt} > \Gamma_{null}$, then ``at least as extreme'' means $\Gamma(s_{null}) \geq \Gamma(s_{alt})$. Otherwise, it means $\Gamma(s_{null}) < \Gamma(s_{alt})$. The estimator is fast to compute if the $\Gamma$ values are presorted, and because it is averaged over a large number of samples from the alternative hypothesis, it is a robust estimator for determining whether $N \times K$ is a large enough sample size.

Finally, as is typical for NHST, we reject the null hypothesis when the $p$-value is below the significance level $\alpha=0.05$.


\section{Results on Additional Datasets}
\label{appendix:additional_results}

The \textbf{Amazon reviews} dataset \cite{zhang2015character} contains 20,415 products rated by 5 reviewers on a scale of 1-5, which were selected from the full dataset of reviews from $6,643,669$ users on $2,441,053$ products from those products having at least 5 reviews. We fit the means and standard deviations of the item responses to \emph{truncated} normal distributions with ($\mu=0.552121, \sigma=0.032093)$ and $(\mu=0.318177, \sigma=0.018281)$, respectively.

The \textbf{HS-Brexit} dataset \cite{akhtar2021whose} contains 1120 tweets related to Brexit and is labeled for hate speech by 6 raters each. We fit the means and standard deviations of the item responses to \emph{truncated} normal distributions with ($\mu=-0.278260, \sigma=0.181938)$ and $(\mu=-0.340141, \sigma=0.408186)$, respectively.

The \textbf{ConvAbuse} dataset \cite{cercas-curry-etal-2021-convabuse} contains 4185 dialogues between users and two conversational agents and is labeled for abuse by at least 3 experts each. We fit the means and standard deviations of the item responses to \emph{truncated} normal distributions with ($\mu=1.124694, \sigma=0.512993)$ and $(\mu=-0.324344, \sigma=0.417337)$, respectively.

The \textbf{ArMIS} dataset \cite{almanea-poesio-2022-armis} contains 964 Arabic tweets for misogyny detection and is labeled by 3 raters each. We fit the means and standard deviations of the item responses to \emph{truncated} bi-normal distributions with ($\mu_1=-0.430701, \sigma_1=0.418148, \mu_2=1.194010, \sigma_1=0.525248)$ with the likelihood of choosing the first distribution as $0.652561$ and $(\mu_1=-0.264113, \sigma_1=0.530150, \mu_2=0.362404, \sigma_2=0.632262)$ with the likelihood of choosing the first distribution as $0.76639$, respectively.

The \textbf{Measuring Hate Speech (MHS)} dataset \cite{sachdeva-etal-2022-measuring} contains 39,565 comments labeled for hate speech by 7912 raters. We fit the means and standard deviations of the item responses to \emph{truncated} normal distributions with ($\mu=-0.211147, \sigma=0.106442)$ and $(\mu=-0.243672, \sigma=0.148406)$, respectively.

\begin{figure*}
  \centering
  \begin{subfigure}[b]{0.3\linewidth}
    \centering
    \includegraphics[width=\linewidth]{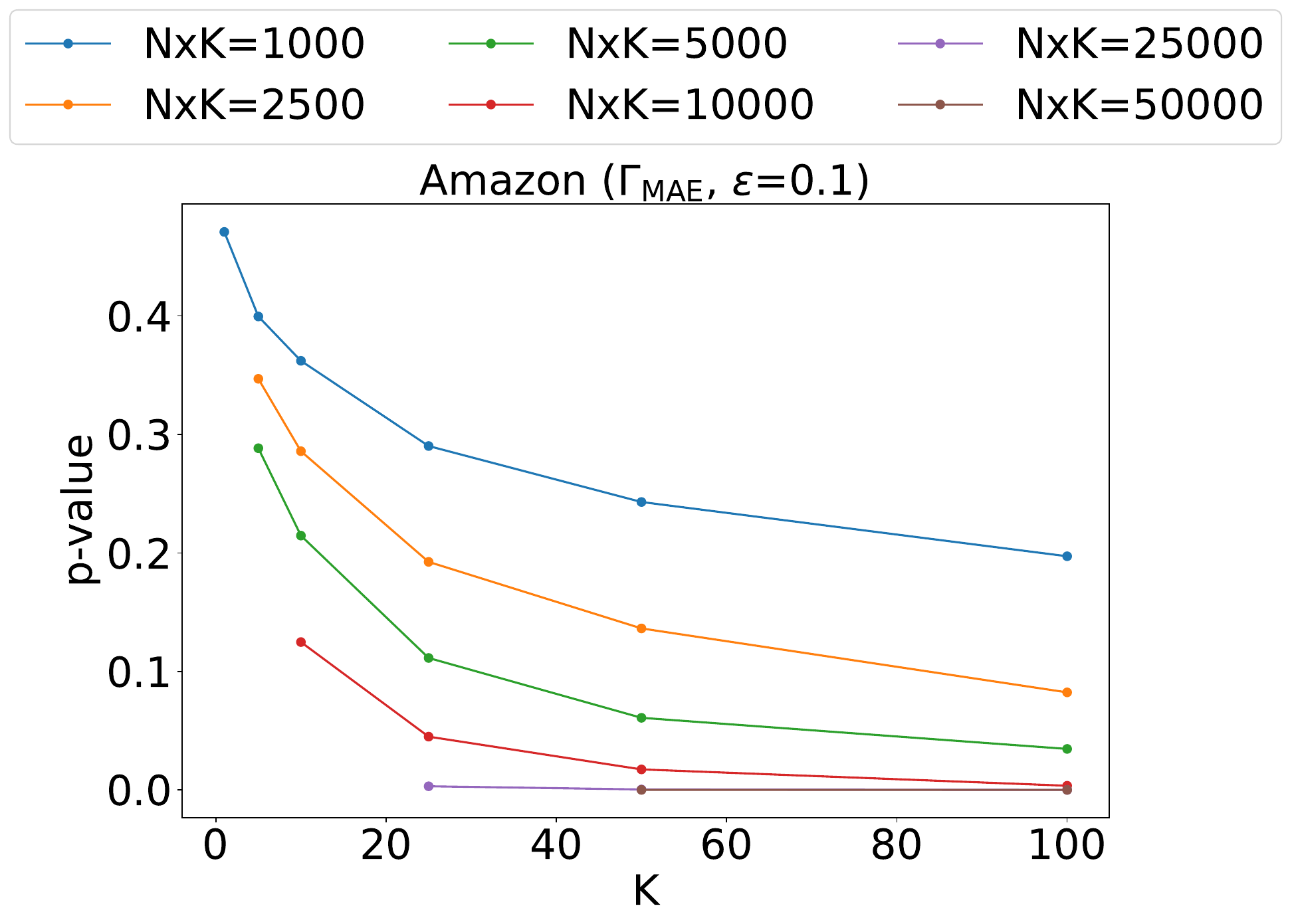}
    \caption{MAE}
    \label{fig:amazon_mae_e01}
  \end{subfigure} \hfill
  \begin{subfigure}[b]{0.3\linewidth}
    \centering
    \includegraphics[width=\linewidth]{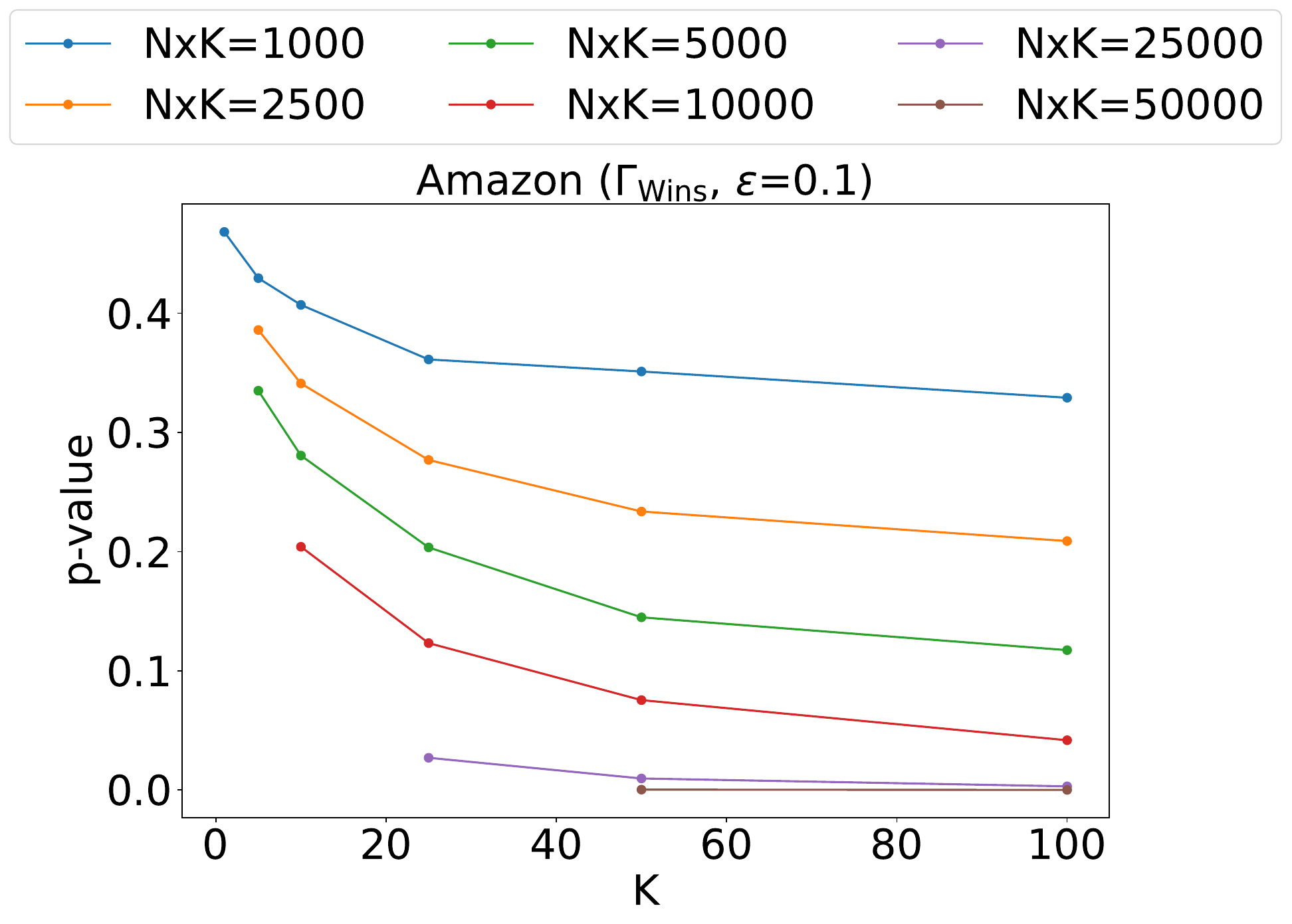}
    \caption{Wins}
    \label{fig:amazon_wins_e01}
  \end{subfigure} \hfill
  \begin{subfigure}[b]{0.3\linewidth}
    \centering
    \includegraphics[width=\linewidth]{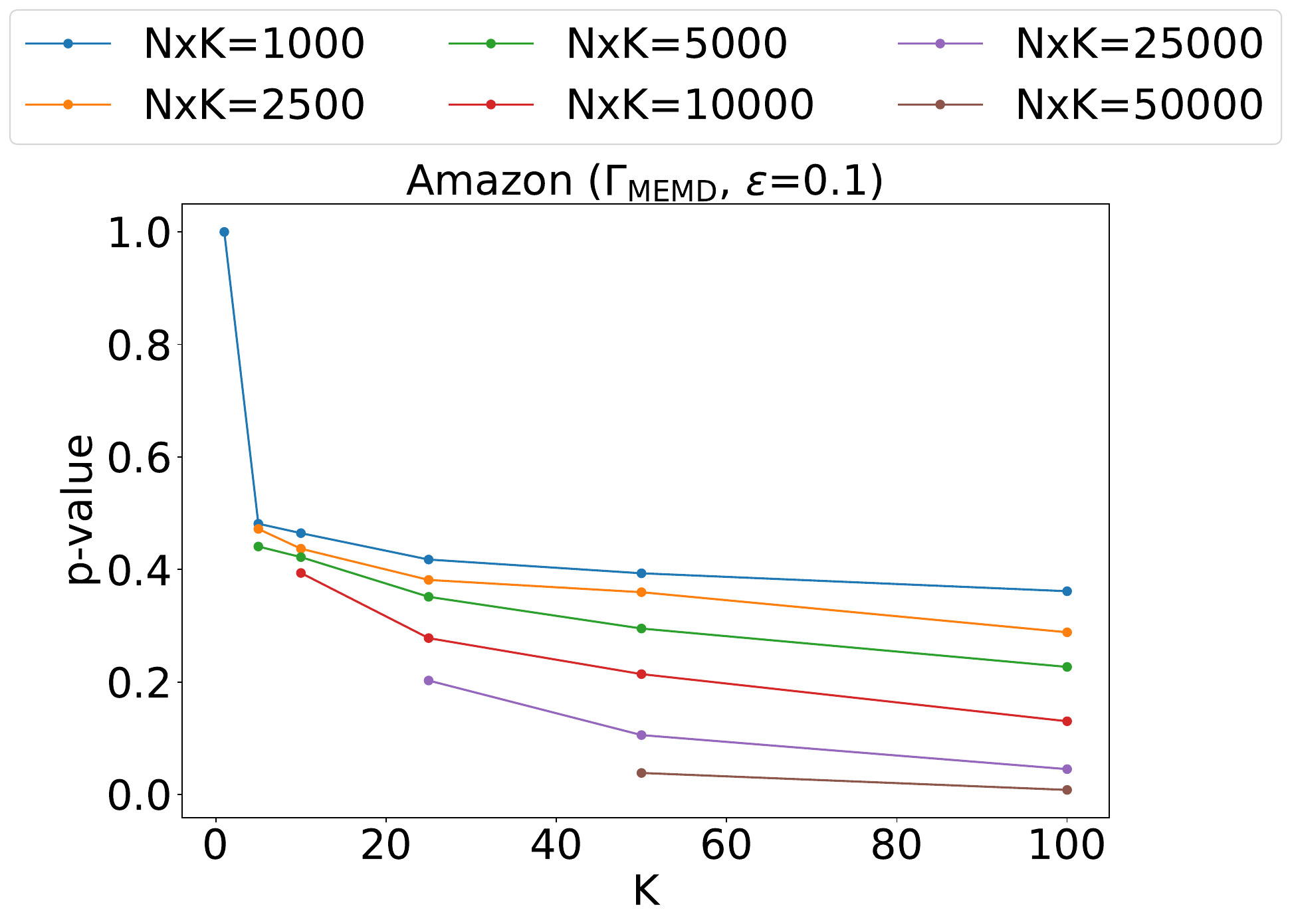}
    \caption{MEMD}
    \label{fig:amazon_memd_e03}
  \end{subfigure}
  \caption{\pv\ plots for Amazon dataset, $\epsilon=0.1$.}
  \label{fig:amazon_p_vals_e01}
\end{figure*}

\begin{figure*}
  \centering
  \begin{subfigure}[b]{0.3\linewidth}
    \centering
    \includegraphics[width=\linewidth]{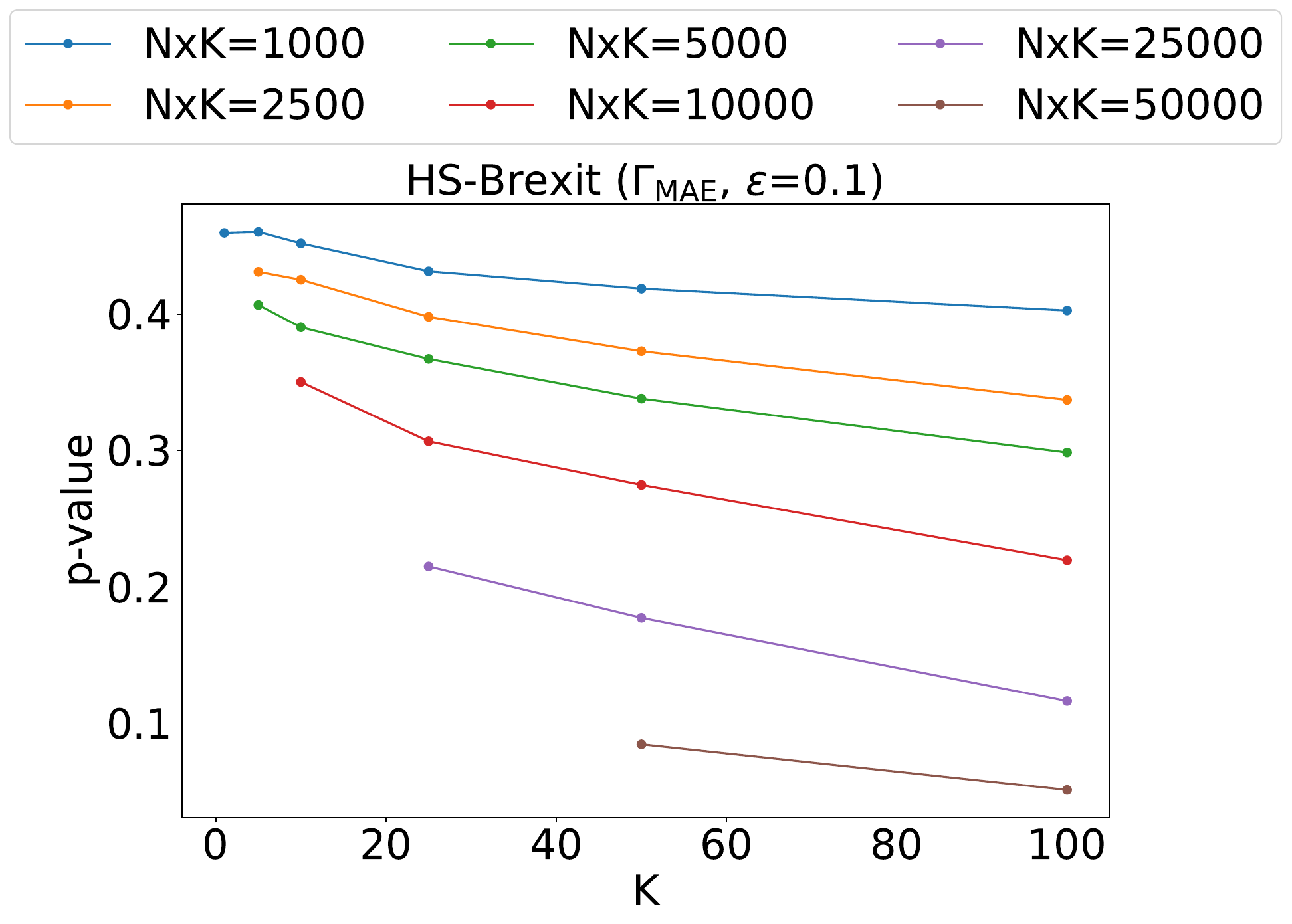}
    \caption{MAE}
    \label{fig:hs_brexit_mae_e01}
  \end{subfigure} \hfill
  \begin{subfigure}[b]{0.3\linewidth}
    \centering
    \includegraphics[width=\linewidth]{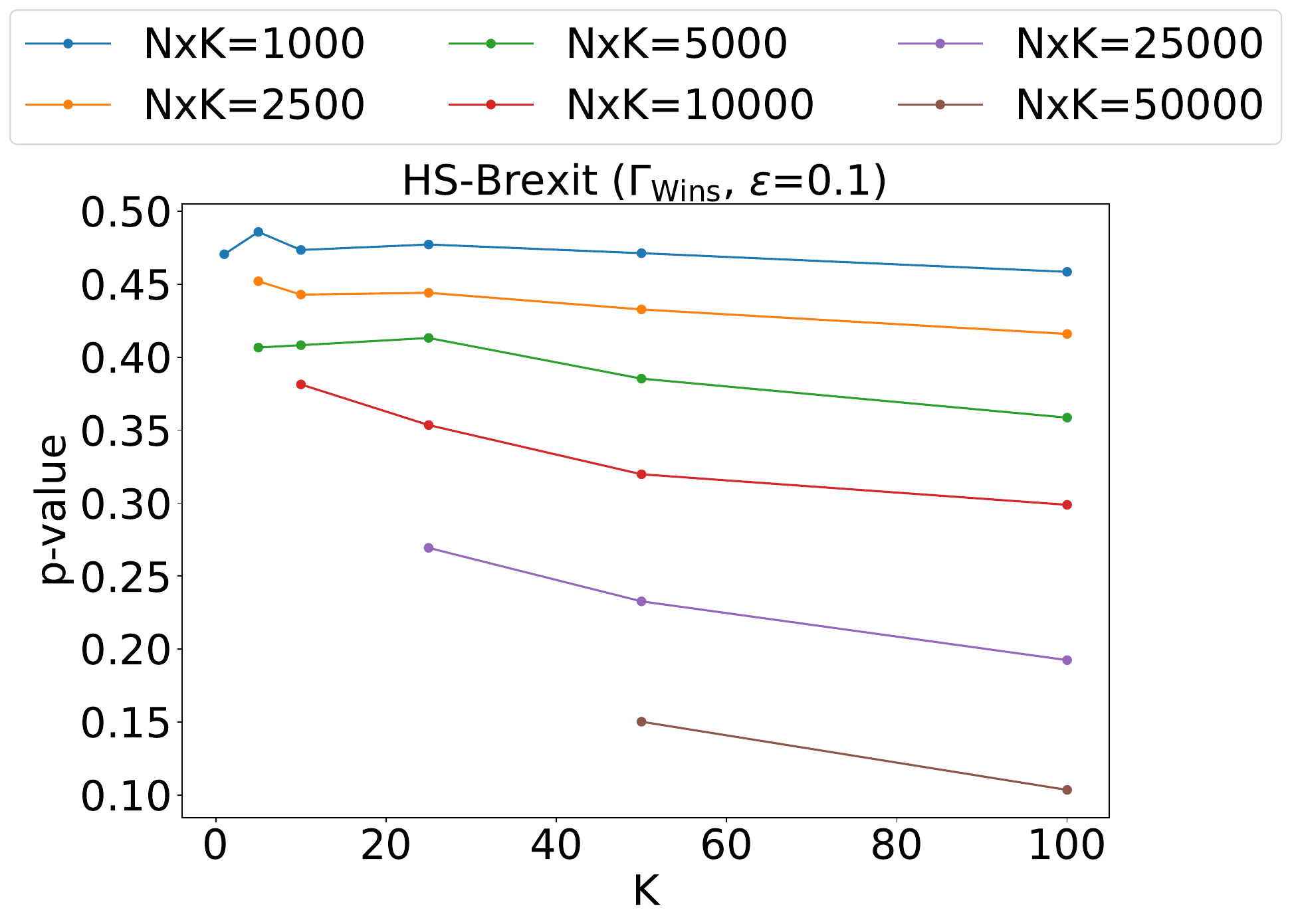}
    \caption{Wins}
    \label{fig:hs_brexit_wins_e01}
  \end{subfigure} \hfill
  \begin{subfigure}[b]{0.3\linewidth}
    \centering
    \includegraphics[width=\linewidth]{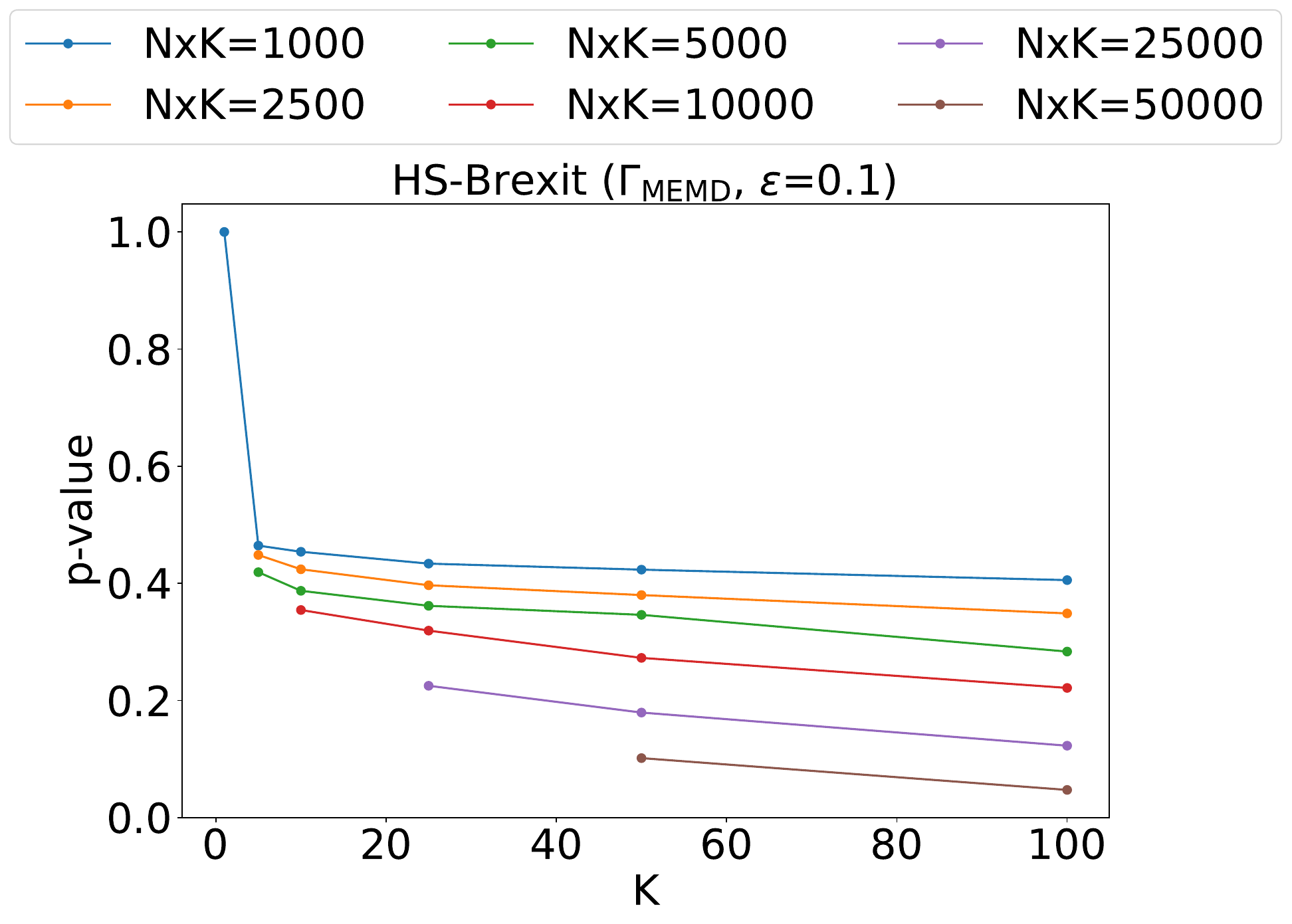}
    \caption{MEMD}
    \label{fig:hs_brexit_memd_e03}
  \end{subfigure}
  \caption{\pv\ plots for HS-Brexit dataset, $\epsilon=0.1$.}
  \label{fig:hs_brexit_p_vals_e01}
\end{figure*}

\begin{figure*}
  \centering
  \begin{subfigure}[b]{0.3\linewidth}
    \centering
    \includegraphics[width=\linewidth]{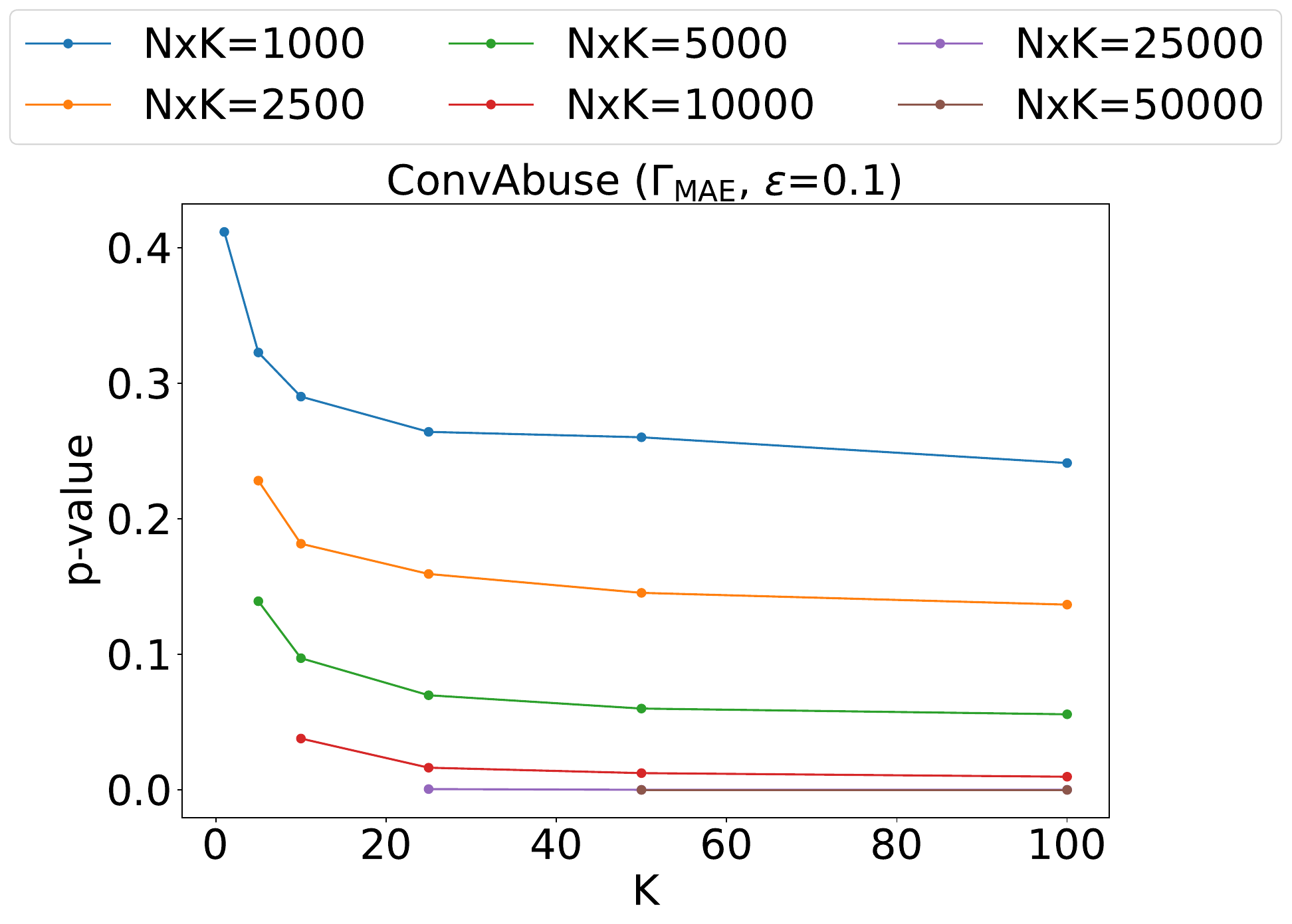}
    \caption{MAE}
    \label{fig:convabuse_mae_e01}
  \end{subfigure} \hfill
  \begin{subfigure}[b]{0.3\linewidth}
    \centering
    \includegraphics[width=\linewidth]{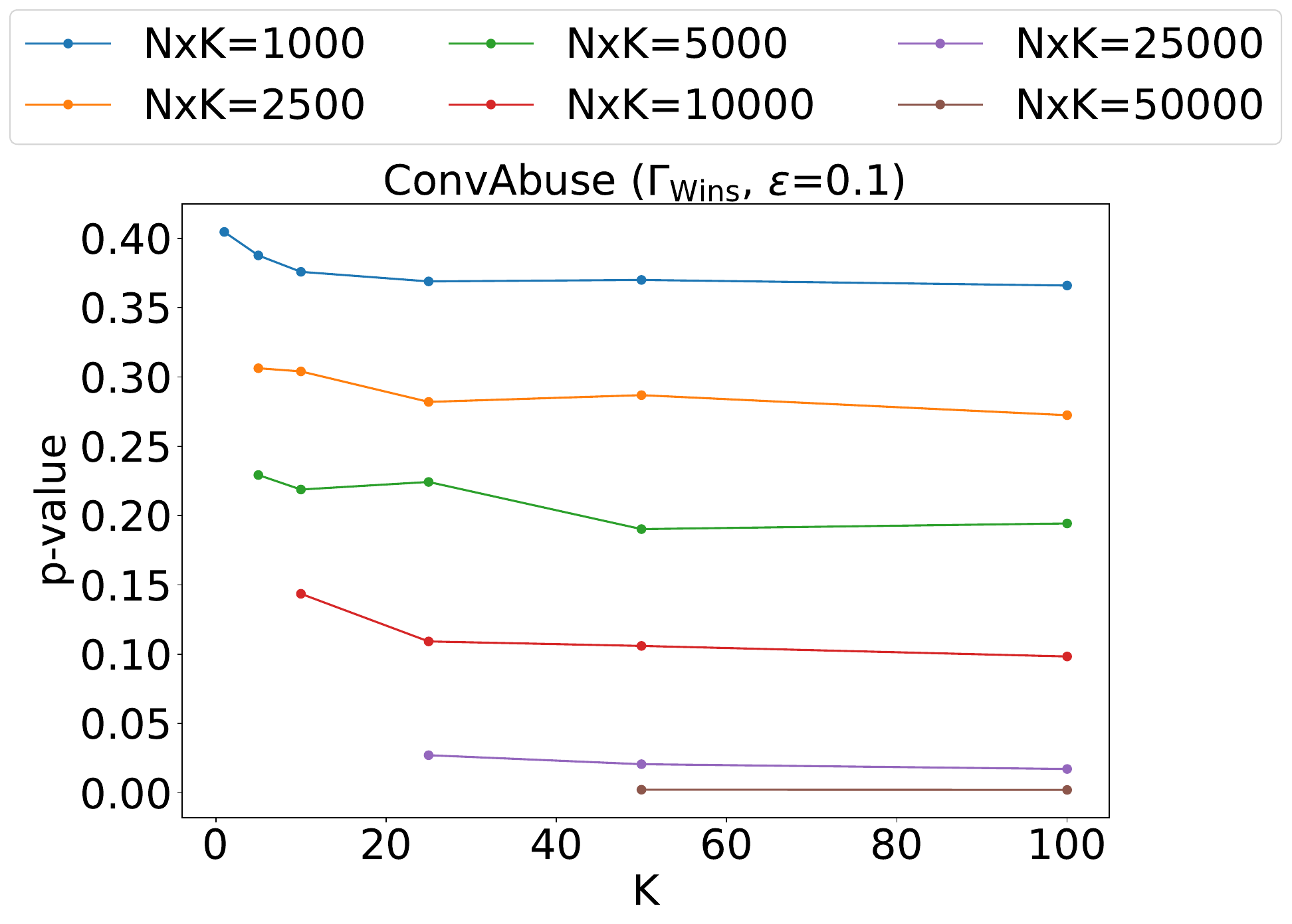}
    \caption{Wins}
    \label{fig:convabuse_wins_e01}
  \end{subfigure} \hfill
  \begin{subfigure}[b]{0.3\linewidth}
    \centering
    \includegraphics[width=\linewidth]{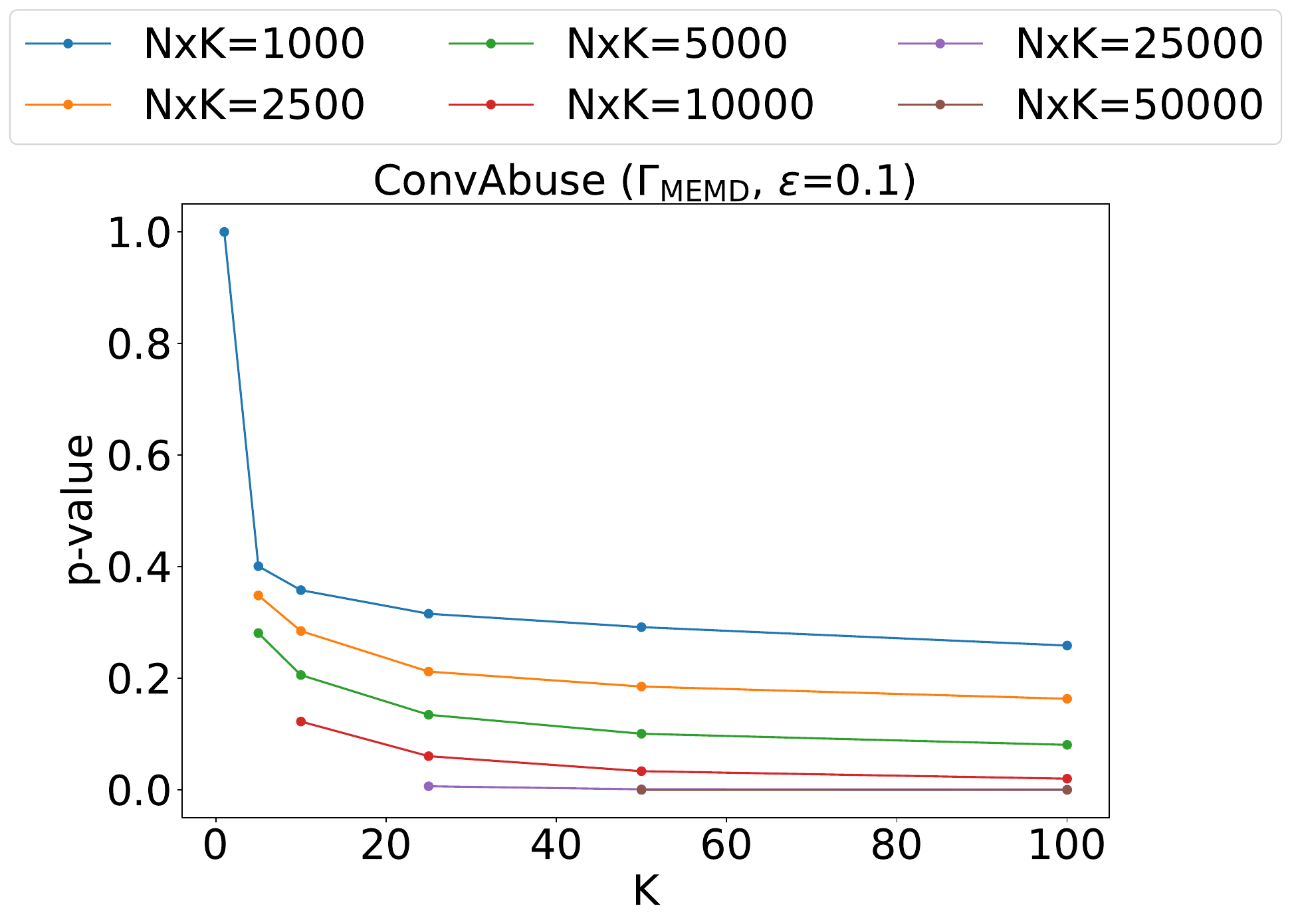}
    \caption{MEMD}
    \label{fig:convabuse_memd_e03}
  \end{subfigure}
  \caption{\pv\ plots for ConvAbuse dataset, $\epsilon=0.1$.}
  \label{fig:convabuse_p_vals_e01}
\end{figure*}

\begin{figure*}
  \centering
  \begin{subfigure}[b]{0.3\linewidth}
    \centering
    \includegraphics[width=\linewidth]{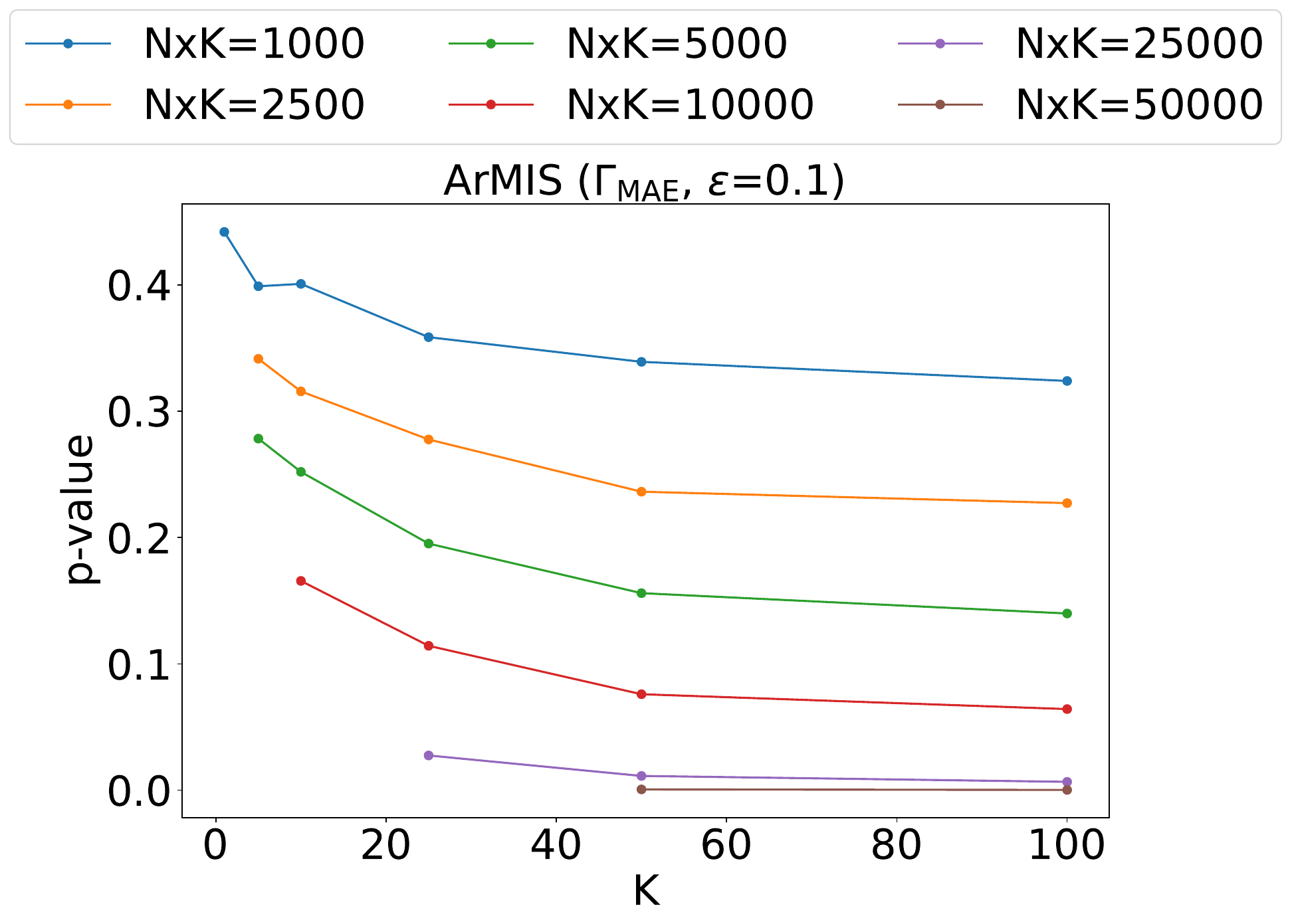}
    \caption{MAE}
    \label{fig:armis_mae_e01}
  \end{subfigure} \hfill
  \begin{subfigure}[b]{0.3\linewidth}
    \centering
    \includegraphics[width=\linewidth]{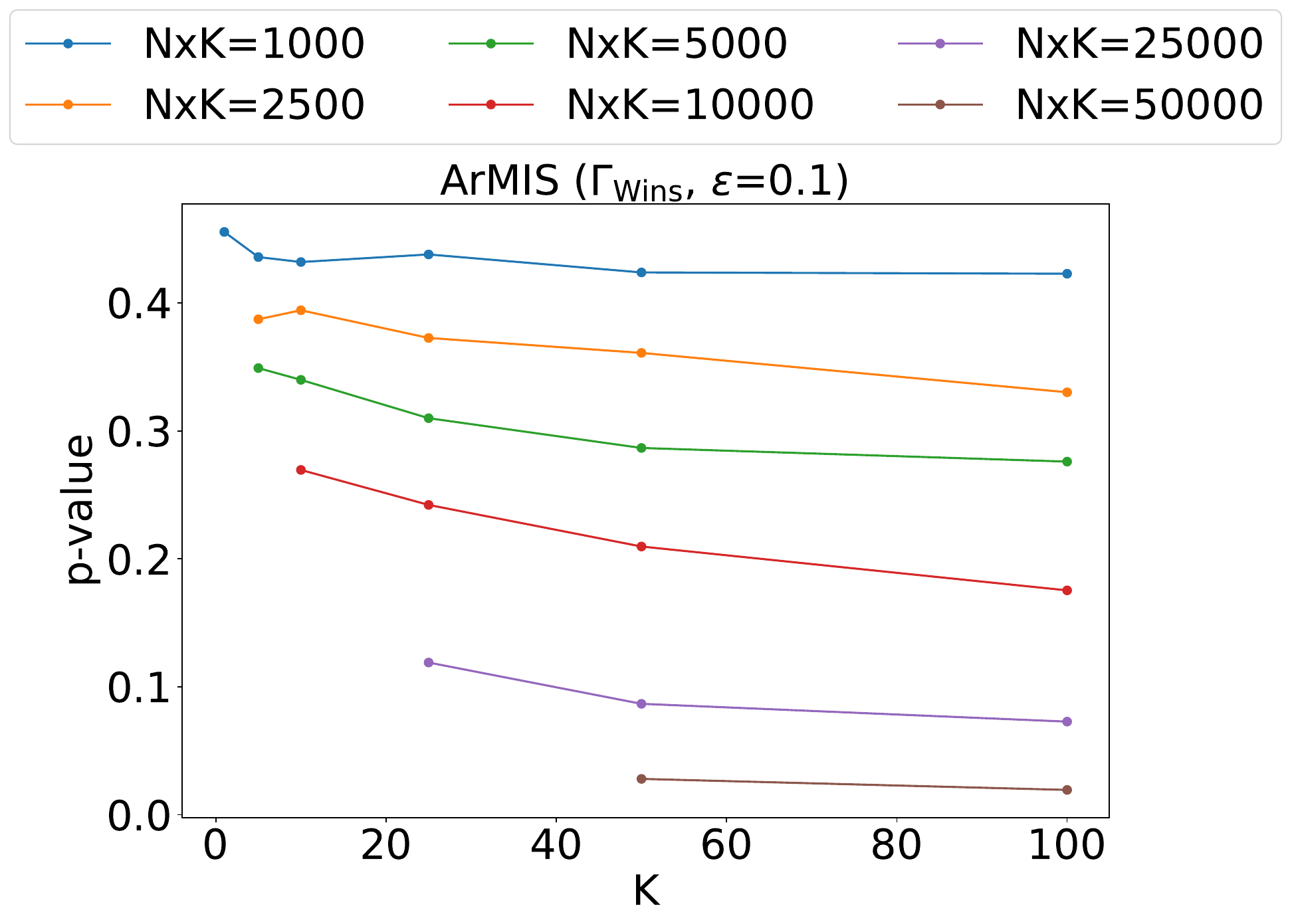}
    \caption{Wins}
    \label{fig:armis_wins_e01}
  \end{subfigure} \hfill
  \begin{subfigure}[b]{0.3\linewidth}
    \centering
    \includegraphics[width=\linewidth]{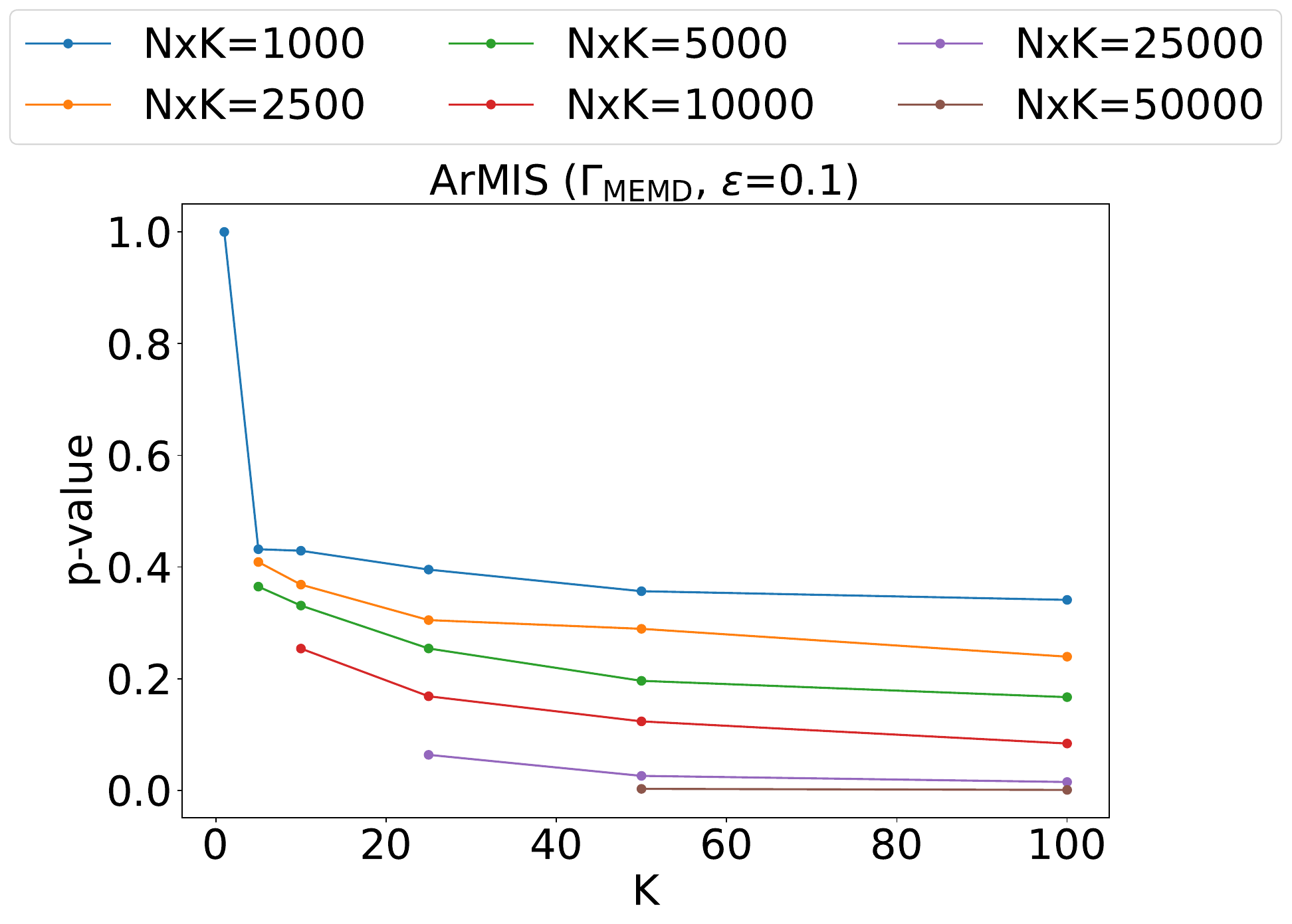}
    \caption{MEMD}
    \label{fig:armis_memd_e03}
  \end{subfigure}
  \caption{\pv\ plots for ArMIS dataset, $\epsilon=0.1$.}
  \label{fig:armis_p_vals_e01}
\end{figure*}

\begin{figure*}
  \centering
  \begin{subfigure}[b]{0.3\linewidth}
    \centering
    \includegraphics[width=\linewidth]{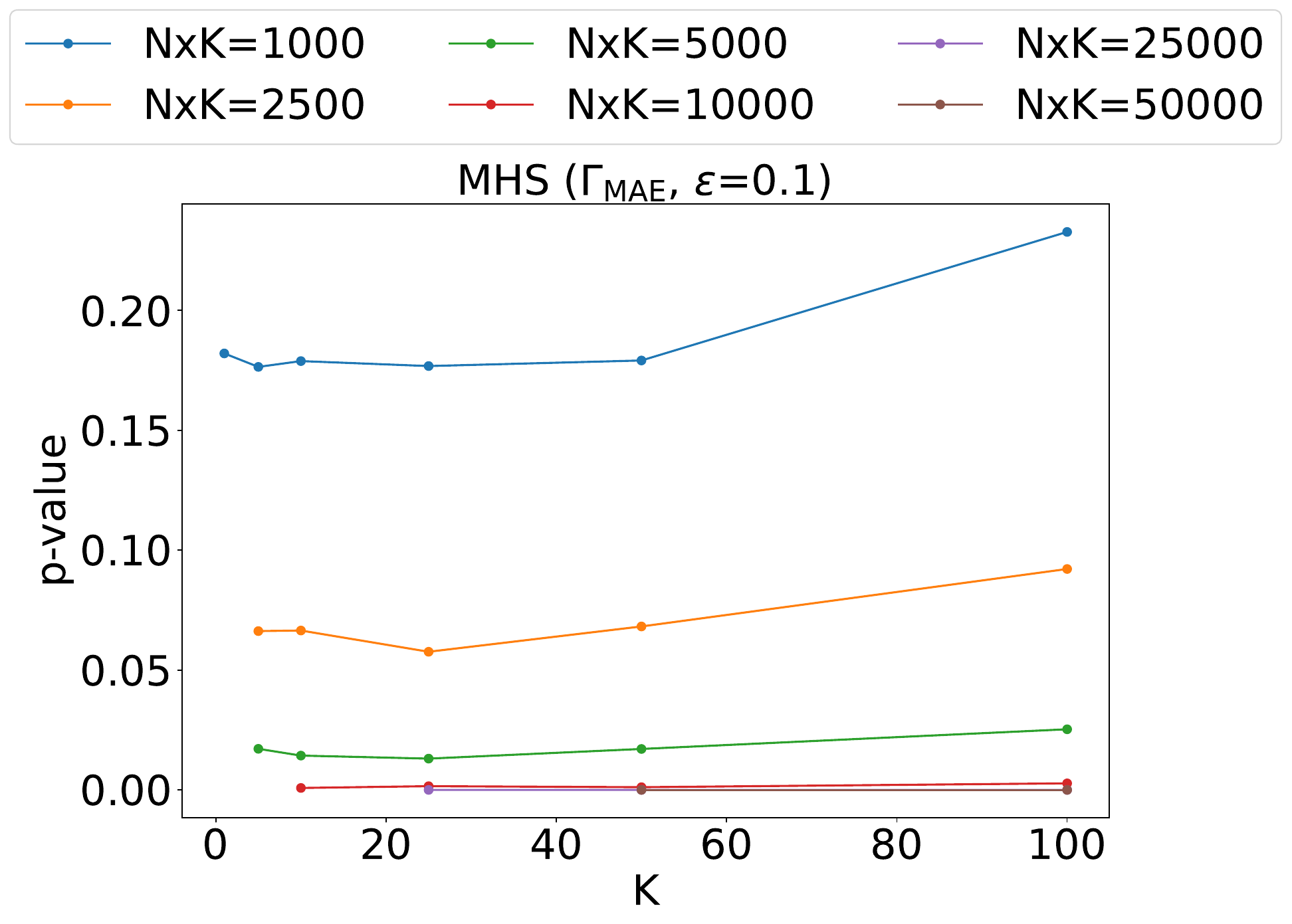}
    \caption{MAE}
    \label{fig:mhs_mae_e01}
  \end{subfigure} \hfill
  \begin{subfigure}[b]{0.3\linewidth}
    \centering
    \includegraphics[width=\linewidth]{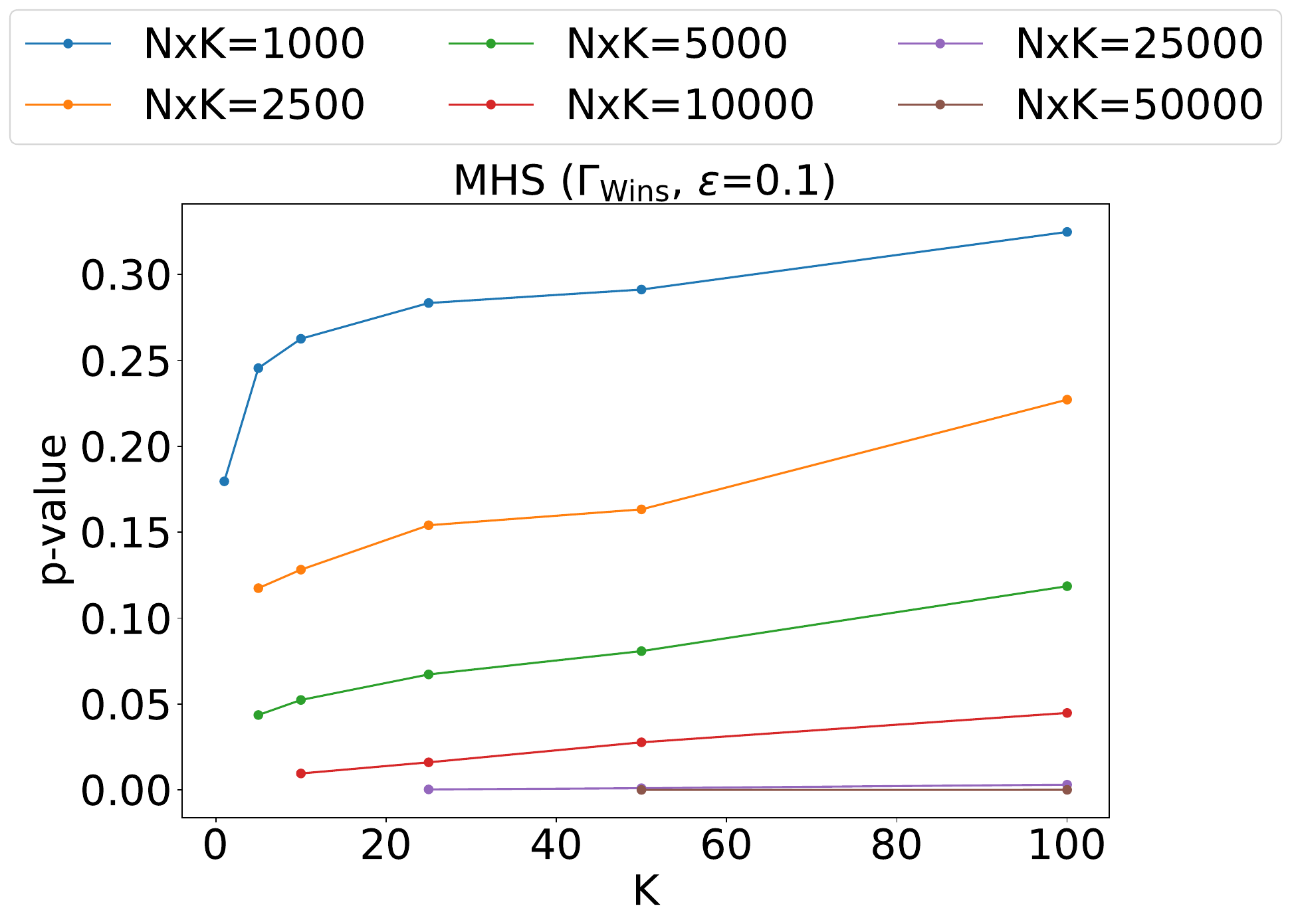}
    \caption{Wins}
    \label{fig:mhs_wins_e01}
  \end{subfigure} \hfill
  \begin{subfigure}[b]{0.3\linewidth}
    \centering
    \includegraphics[width=\linewidth]{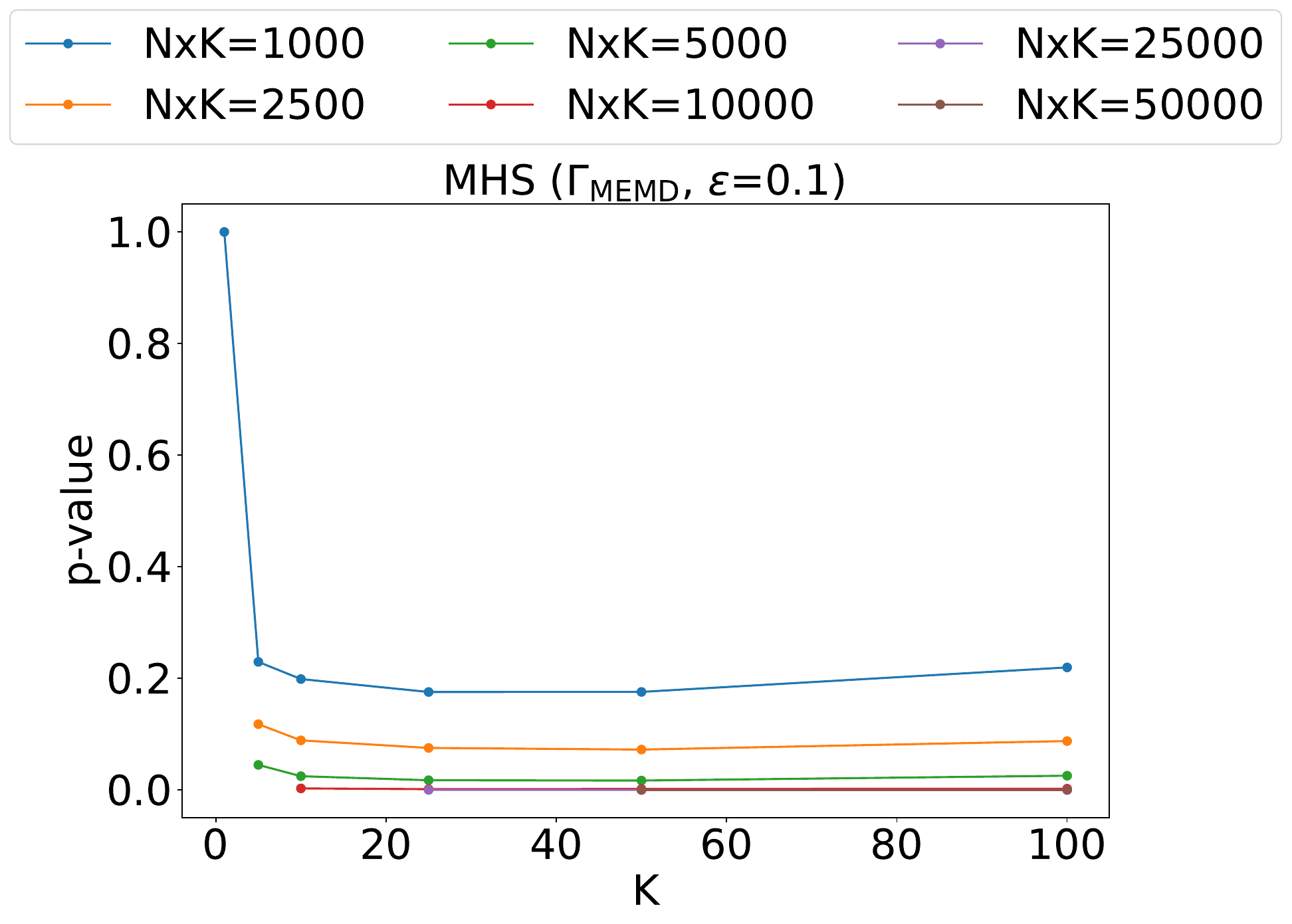}
    \caption{MEMD}
    \label{fig:mhs_memd_e03}
  \end{subfigure}
  \caption{\pv\ plots for MHS dataset, $\epsilon=0.1$.}
  \label{fig:mhs_p_vals_e01}
\end{figure*}

Tables \ref{tab:low_k_for_p_lt_05_nk_min_p_dist_0.05}--\ref{tab:low_k_for_p_lt_05_nk_min_p_dist_0.2} show the results for minimum \pv, $K$, and corresponding effect size ($\Delta$) for lowest $NK$ with $p<0.05$ for different $ \epsilon$. In Table \ref{tab:low_k_for_p_lt_05_nk_min_p_dist_0.1} ($\epsilon=0.1$), we observe that minimum \pv s are consistently obtained with a higher $K=100$ for all datasets except MHS, where minimum \pv s are obtained at $K=\{5,10\}$. We notice a similar trend for $\epsilon=0.0.05$ and $\epsilon=0.2$. Figures \ref{fig:amazon_p_vals_e01}--\ref{fig:mhs_p_vals_e01} show results for \pv s for $\epsilon = 0.1$ for different datasets and metric combinations.

\begin{table}[h]
\centering
\small
\begin{tabular}{l|c|ccc}
Dataset & Stat & $\Gamma_{\rm MAE}$ & $\Gamma_{\rm Wins}$ & $\Gamma_{\rm MEMD}$ \\
\midrule
 & NK & 10000 & 25000 & 25000 \\
Toxicity & p-value & 0.047 & 0.026 & 0.011 \\
 & K & 100 & 100 & 100 \\
 & $\Delta$ & 0.007 & 0.227 & 0.546 \\
 \hline
 & NK & 100000 & - & 100000 \\
MultiDomain & p-value & 0.019 & - & 0.033 \\
 & K & 100 & - & 100 \\
 & $\Delta$ & 0.006 & - & 0.195 \\
 \hline
 & NK & 40000 & 100000 & - \\
Amazon & p-value & 0.043 & 0.031 & - \\
 & K & 100 & 100 & - \\
 & $\Delta$ & 0.005 & 0.118 & - \\
 \hline
 & NK & - & - & - \\
HS-Brexit & p-value & - & - & - \\
 & K & - & - & - \\
 & $\Delta$ & - & - & - \\
 \hline
 & NK & 40000 & 100000 & 40000 \\
ConvAbuse & p-value & 0.020 & 0.037 & 0.033 \\
 & K & 100 & 100 & 100 \\
 & $\Delta$ & 0.009 & 0.093 & 0.279 \\
 \hline
 & NK & 100000 & - & 100000 \\
ArMIS & p-value & 0.025 & - & 0.047 \\
 & K & 100 & - & 100 \\
 & $\Delta$ & 0.005 & - & 0.169 \\
 \hline
 & NK & 20000 & 40000 & 20000 \\
MHS & p-value & 0.041 & 0.044 & 0.037 \\
 & K & 100 & 100 & 100 \\
 & $\Delta$ & 0.001 & 0.107 & 0.246 \\
\end{tabular}
\caption{Minimum \pv, $K$, and corresponding effect size ($\Delta$) for lowest $NK$ with $p<0.05$ ($ \epsilon=0.05$).}
\label{tab:low_k_for_p_lt_05_nk_min_p_dist_0.05}
\end{table}

\begin{table}[h]
\centering
\small
\begin{tabular}{l|c|ccc}
Dataset & Stat & $\Gamma_{\rm MAE}$ & $\Gamma_{\rm Wins}$ & $\Gamma_{\rm MEMD}$ \\
\midrule
 & NK & 2000 & 5000 & 4000 \\
Toxicity & p-value & 0.041 & 0.040 & 0.024 \\
 & K & 100 & 100 & 100 \\
 & $\Delta$ & 0.021 & 0.443 & 1.403 \\
 \hline
 & NK & 20000 & 40000 & 20000 \\
MultiDomain & p-value & 0.015 & 0.030 & 0.024 \\
 & K & 100 & 100 & 100 \\
 & $\Delta$ & 0.018 & 0.169 & 0.546 \\
 \hline
 & NK & 4000 & 10000 & 25000 \\
Amazon & p-value & 0.039 & 0.042 & 0.045 \\
 & K & 100 & 100 & 100 \\
 & $\Delta$ & 0.018 & 0.335 & 0.331 \\
 \hline
 & NK & 100000 & 100000 & 50000 \\
HS-Brexit & p-value & 0.009 & 0.037 & 0.047 \\
 & K & 100 & 100 & 100 \\
 & $\Delta$ & 0.004 & 0.077 & 0.153 \\
 \hline
 & NK & 10000 & 20000 & 10000 \\
ConvAbuse & p-value & 0.010 & 0.028 & 0.020 \\
 & K & 100 & 100 & 100 \\
 & $\Delta$ & 0.025 & 0.212 & 0.732 \\
 \hline
 & NK & 20000 & 40000 & 20000 \\
ArMIS & p-value & 0.018 & 0.035 & 0.024 \\
 & K & 100 & 100 & 100 \\
 & $\Delta$ & 0.016 & 0.158 & 0.491 \\
 \hline
 & NK & 4000 & 5000 & 4000 \\
MHS & p-value & 0.028 & 0.044 & 0.040 \\
 & K & 10 & 5 & 10 \\
 & $\Delta$ & 0.004 & 0.049 & 0.053 \\
\end{tabular}
\caption{Minimum \pv, $K$, and corresponding effect size ($\Delta$) for lowest $NK$ with $p<0.05$ ($ \epsilon=0.1$).}
\label{tab:low_k_for_p_lt_05_nk_min_p_dist_0.1}
\end{table}

\begin{table}[h]
\centering
\small
\begin{tabular}{l|c|ccc}
Dataset & Stat & $\Gamma_{\rm MAE}$ & $\Gamma_{\rm Wins}$ & $\Gamma_{\rm MEMD}$ \\
\midrule
 & NK & 500 & 1250 & 1250 \\
Toxicity & p-value & 0.048 & 0.045 & 0.046 \\
 & K & 50 & 5 & 50 \\
 & $\Delta$ & 0.045 & 0.192 & 1.056 \\
 \hline
 & NK & 2500 & 10000 & 4000 \\
MultiDomain & p-value & 0.042 & 0.026 & 0.029 \\
 & K & 50 & 50 & 100 \\
 & $\Delta$ & 0.042 & 0.246 & 1.390 \\
 \hline
 & NK & 1000 & 2500 & 4000 \\
Amazon & p-value & 0.024 & 0.041 & 0.018 \\
 & K & 100 & 50 & 100 \\
 & $\Delta$ & 0.053 & 0.479 & 1.220 \\
 \hline
 & NK & 10000 & 20000 & 10000 \\
HS-Brexit & p-value & 0.024 & 0.030 & 0.024 \\
 & K & 100 & 50 & 100 \\
 & $\Delta$ & 0.014 & 0.129 & 0.531 \\
 \hline
 & NK & 2000 & 4000 & 2000 \\
ConvAbuse & p-value & 0.023 & 0.037 & 0.044 \\
 & K & 50 & 10 & 100 \\
 & $\Delta$ & 0.057 & 0.127 & 1.659 \\
 \hline
 & NK & 4000 & 10000 & 4000 \\
ArMIS & p-value & 0.018 & 0.028 & 0.034 \\
 & K & 100 & 25 & 100 \\
 & $\Delta$ & 0.043 & 0.154 & 1.250 \\
 \hline
  & NK & 200 & 200 & 1000 \\
MHS & p-value & 0.038 & 0.039 & 0.016 \\
 & K & 1 & 1 & 10 \\
 & $\Delta$ & 0.020 & 0.097 & 0.185 \\
\end{tabular}
\caption{Minimum \pv, $K$, and corresponding effect size ($\Delta$) for lowest $NK$ with $p<0.05$ ($ \epsilon=0.2$).}
\label{tab:low_k_for_p_lt_05_nk_min_p_dist_0.2}
\end{table}


\hide{
\subsection{Power Analysis}
\begin{figure}[htb]
\begin{subfigure}[]{0.5\textwidth}
\includegraphics[width=2.5in]{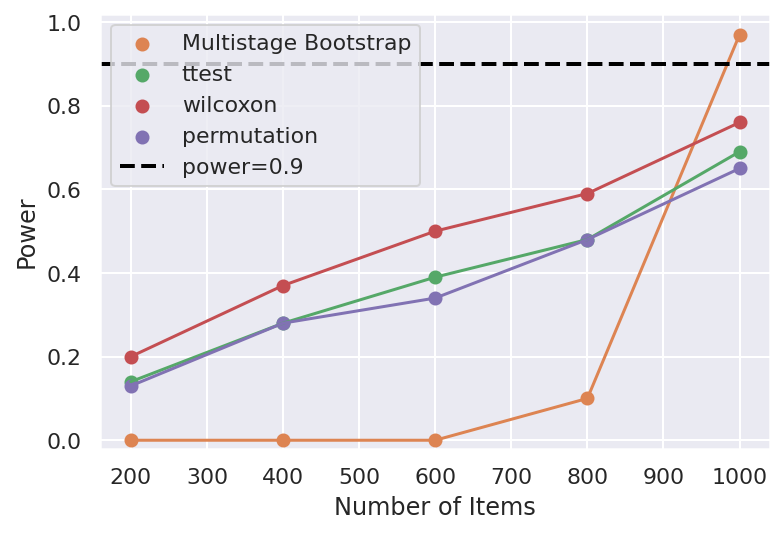}
\caption{Varying $N$ with $K=5$}
\end{subfigure}
\begin{subfigure}[]{0.5\textwidth}
\includegraphics[width=2.5in]{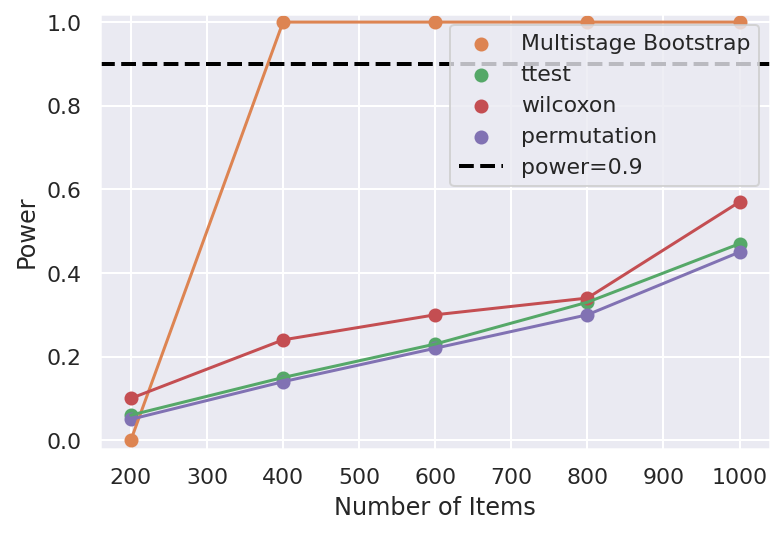}
\caption{Varying $N$ with $K=10$}
\end{subfigure}
\begin{subfigure}[]{0.5\textwidth}
\includegraphics[width=2.5in]{images/md_power_analysis_N=1000_eps=0_1.png}
\caption{Varying $K$ with $N=1000$}
\end{subfigure}
\caption{Power Analysis of MultiDomain data ($\epsilon=0.1)$. Each data point is the estimated from 1000 outer-level samples, each consisting of 10000 inner level samples.}
\label{fig:power_vs_sample_size}
\end{figure}
}

\end{document}